\documentclass[11pt]{article}

\usepackage[preprint]{acl}
\usepackage{times}
\usepackage{latexsym}

\usepackage[T1]{fontenc}
\usepackage[utf8]{inputenc}

\usepackage{microtype}

\usepackage{inconsolata}

\usepackage{graphicx}

\usepackage{multirow}
\usepackage{amsmath,amsfonts,amssymb}
\usepackage{pgf}

\usepackage{algorithm}
\usepackage[noend]{algpseudocode}
\algnewcommand\algorithmicforeach{\textbf{for each}}
\algdef{S}[FOR]{ForEach}[1]{\algorithmicforeach\ #1\ \algorithmicdo}

\usepackage{enumerate}
\usepackage[shortlabels]{enumitem}

\usepackage[dvipsnames]{xcolor}

\definecolor{BASE}{HTML}{FFA500}
\definecolor{DIVERSE}{HTML}{0072B2}
\definecolor{FULL}{HTML}{332288}
\definecolor{RANDOM}{HTML}{8C564B}

\definecolor{EX}{HTML}{66CC77}

\newcommand{\BASE}[1]{\textcolor{BASE}{#1}}
\newcommand{\DIVERSE}[1]{\textcolor{DIVERSE}{#1}}
\newcommand{\RANDOM}[1]{\textcolor{RANDOM}{#1}}
\newcommand{\FULL}[1]{\textcolor{FULL}{#1}}

\newcommand{\EX}[1]{\textcolor{EX}{#1}}

\def\true{1}

\newcommand{\longpaperinput}[1]{\if\islongpaper\true\input{#1}\fi}

\providecommand{\tightlist}{%
 \setlength{\itemsep}{0pt}%
 \setlength{\parskip}{0pt}%
 }

\newcommand{\PERFORMANCEPATH}[1]{}

\title{A Diversity Diet for a Healthier Model: A Case Study of French ModernBERT}

\author{%
  Louis Estève \\
  Université Paris-Saclay, CNRS, LISN \\
  France \\
  \texttt{first.last@lisn.fr} \\\And
  Christophe Servan \\
  Université Paris-Saclay, CNRS, LISN \\
  AMIAD, Pôle Recherche \\
  France \\
  \texttt{first.last@polytechnique.edu} \\ \AND
  Thomas Lavergne \\
  Université Paris-Saclay, CNRS, LISN \\
  France \\
  \texttt{first.last@lisn.fr} \\\And
  Agata Savary \\
  Université Paris-Saclay, CNRS, LISN \\
  France \\
  \texttt{first.last@lisn.fr} \\ModernBERT
}

\author{
  \textbf{Louis Estève\textsuperscript{1}},
  \textbf{Christophe Servan\textsuperscript{1,2}},
  \textbf{Thomas Lavergne\textsuperscript{1}},
  \textbf{Agata Savary\textsuperscript{1}}
\vspace*{0.1cm}
\\ 
  \textsuperscript{1} Université Paris-Saclay, CNRS, LISN, \texttt{first.last@lisn.fr}
  \vspace*{0.1cm}
  \\
  \textsuperscript{2}AMIAD, Pôle Recherche, \texttt{first.last@polytechnique.edu}
}

\begin{document}
\maketitle
  \begin{abstract}
  Diversity has been gaining interest in the NLP community in recent years. 
  At the same time, state-of-the-art transformer models such as ModernBERT use very
  large pre-training datasets, which are driven by size rather than by
  diversity. This summons to investigate %
  the impact of diversity on %
  pre-training. We do so in this study, with the express intent of reducing pre-training dataset size, while retaining at least comparable
  performance. We compare diversity-driven sampling algorithms, and we use the best one to pre-train several ModernBERT models on French with a fixed compute budget. We fine-tune and evaluate them on a variety of French benchmarks. We compare them with models pre-trained on randomly sampled data of commensurate size, with the same compute budget.  We find that both random and diversity-driven sampling may reduce the pre-training dataset by up to 94\% and the pre-training time by up to 73\% while maintaining performance.  Moreover, in some tasks, the inherent quality of models, estimated via head-only fine-tuning, is up to 10 points higher with diversity sampling than with random sampling.

\end{abstract}

\section{Introduction}

Natural Language Processing (NLP) has seen over the years a substantial
increase in dataset size. Datasets ten years ago,
e.g. Universal Dependencies v1.0 \citep{nivre_universal_2016},
had millions of tokens,
unlike some modern datasets
comprising over ten trillion tokens, such as FineWeb \citep{penedo_fineweb_2024} and
HPLT \citep{de-gibert-etal-2024-new-massive}.\footnote{A million is $10^{6}$,
while a trillion is $10^{12}$.} Scalable architectures such as transformers
\citep{vaswani_attention_2017} benefitted from this growth as a consequence of
neural scaling laws. These laws state that, overall, more weights, training
computation, and training data improves model performance
\citep{kaplan2020scalinglawsneurallanguage}. %
This however comes with an exorbitant cost: 
managing vast datasets, and using them to train Large Language Models (LLMs), is non-trivial,
excessively expensive, and detrimental to the environment
\citep{strubell-etal-2019-energy}.

It might be that LLMs benefit from larger datasets due to increased vocabulary
diversity. However, naturally-occurring redundancy in data implies that
randomly appending new data brings diminishing increases in diversity. In other
words, transformers may not be as data-hungry, if we give them
diverse pre-training data. Our research question is then: can diversity-driven
sampling prevent the overgrowth of pre-training datasets, while
preserving or
increasing performance? To address this question, we
set the following hypotheses:
\begin{enumerate}[label=H\arabic*,noitemsep]
  \tightlist{}
  \item A small dataset, with lexical diversity substantially higher than
    at random, can be sampled efficiently from a large dataset.\label{hyp:h1}
  \item A model trained on a small lexically diverse dataset can be competitive to or outperform one trained on a very large dataset.\label{hyp:h2}

\end{enumerate}

We test these hypotheses by (1) proposing two diversity-driven sampling algorithms: one using a heuristic and one based on gradient descent, (2) benchmarking these two algorithms, along with other algorithms found in the literature, and (3) using the best sampling to pre-train ModernBERT \citep{warner-etal-2025-smarter}
models, and evaluate them after fine-tuning on a variety of tasks in French.\footnote{Models: \url{https://huggingface.co/collections/cservan/french-modernbert} -- Code and scripts: \url{https://gitlab.lisn.upsaclay.fr/esteve/transformers}.}

After presenting previous works (§\ref{sec:previous-works}), the sampling
algorithms we benchmark (§\ref{sec:sampling-algorithms}), and our protocol
(§\ref{sec:experimental-protocol}), we show that increased lexical diversity of
the pre-training data substantially reduces the pre-training dataset size while
maintaining performance. Moreover, for some tasks, we achieve substantially
enhanced performances (§\ref{sec:results}).

\section{Previous works}\label{sec:previous-works}

For the needs of our study, we need to document what is diversity (and how it
is quantified), but also sampling algorithms that relate to our problem, and
the model architecture we shall use.

\subsection{Diversity}\label{sec:diversity}

Diversity has been discussed in multiple fields and may be defined as the study
of categories in a population, through \emph{variety} which focuses on how many
categories are present, \emph{balance} which focuses on how evenly distributed
categories are, and \emph{disparity} which focuses on how fundamentally
different categories are
\citep{stirling_diversity_1994,stirling_power_1994,stirling_general_2007,ramaciotti_morales_measuring_2021}.
Ecology has studied biodiversity as early as the works of
\citet{darwin_descent_1888}, borrowed from information theory
\citep{hill_diversity_1973,patil_diversity_1982}, and further generalised the
concept to build upon functional \citep{chao_unifying_2014} and phylogenetic
\citep{chiu_phylogenetic_2014} distances between species. It is driven by
environmental conversation \citep{sarkar_defining_2002}, and arguably shows maturity in understanding diversity. 
As a consequence, economics has built
upon the works of ecology
\citep{stirling_diversity_1994,stirling_power_1994,stirling_general_2007} and
studied notably the impact of inequalities \citep{ceriani_origins_2012}.
Linguistics has assessed the diversity of languages
\citep{greenberg_measurement_1956,harmon_index_2010}.
NLP as well has recently studied diversity, although with less consensus and
formalisation than in ecology, to the extent that over 100 different measures for
diversity have been observed
\citep{esteve2026surveydiversityquantificationnatural}. Even if focusing on the
diversity of input data, there is no consensus on the measurement of diversity,
as measures can be based on quantity
\citep{bansal-etal-2021-diverse,mohamed-etal-2022-artelingo,bella-etal-2022-language,gueuwou-etal-2023-jwsign,parrish-etal-2024-picture,leeb-scholkopf-2024-diverse,abdelkadir-etal-2024-diverse},
BLEU
\citep{awasthi-etal-2022-diverse,burchell-birch-and-kenneth-heafield-2022-exploring},
distances \citep{stasaski-etal-2020-diverse,shi-etal-2021-diversity},
entropy \citep{stasaski-etal-2020-diverse}, human judgement
\citep{lee-etal-2020-generating}, overlap
\citep{samardzic-etal-2024-measure,yadav-etal-2024-explicit},
type-token-ratio
\citep{liu-etal-2024-prefix,song-etal-2024-scaling,pouran-ben-veyseh-etal-2022-minion},
or distances between distributions \citep{kumar-etal-2022-diversity}.

As a consequence of the lack of consensus and formalisation in NLP, we choose
to measure diversity using entropy \citep{shannon_mathematical_1949,wiener_ergodic_1939}. Our
choice is driven by the fact that the properties of entropy are well-analysed,
and it is a cornerstone for fields such as ecology where the notion of
diversity is well formalised.

\subsection{Diversity-driven sampling}\label{sec:sota-sampling}

Looking at the broad picture of computer science, sampling a dataset for
diversity relates to the knapsack problem \citep{cacchiani_knapsack_2022},
which consists in sampling a set while maximising a metric (and respecting a
size budget). The standard version of this problem requires giving a constant
value (or benefit) to each sentence, which means it can at best be a heuristic
for diversity sampling.

Closer to data found in NLP applications, the Machine Learning literature has a
survey on the topic of data diversity and sampling \citep{gong_diversity_2019}
in which they mend the absence of ``systematical analysis of the
diversification in the machine learning system'' (p.~64323). They present, for
supervised learning, Determinantal Point Processes (DPP) for producing
non-redundant batches out of a fixed dataset, which has the variant $k$-DPP
where $k$ is the requested batch size. DPP has indeed been considered for
enhancing diversity \citep{yang_enhancing_2021,hemmi_lazy_2022}, as avoiding
redundancy may foster diversity, and in turn improve the machine learning
process.

Sampling a dataset with the intent of favouring diversity relates to the
literature on coreset selection, which aims to select ``a subset of the most
informative training samples'' as defined in the surveyby  \cite[p.~181]{guo_deepcore_2022}. A coreset is in
turn more efficient to train on than the full dataset due to reduced size,
while retaining as much performance as possible. Coreset selection has been
used in the perspective of selecting points in a space
\citep{%
  agarwal_geometric_2005,%
  perlitz_active_2023,%
  nguyen_swift_2025%
}, including with the concept of diversity as
pairwise distances \citep{indyk_composable_2014}, which is an instance of
disparity in the cross-domain conceptualisation of diversity
\citep{%
  stirling_general_2007,%
  esteve2026surveydiversityquantificationnatural%
}.
These approaches are at least quadratic to the number of categories time-wise
as they require computing all the distances between categories. Coreset
selection however does not need to rely on distances, and may thus be
subquadratic by focusing on other objects, for instance on confidence, loss or gradient in an already trained model.
Some other approaches consider a specific dataset split to assess which subset of the training split would be a good coreset \citep{san_joaquin_in2core_2024}. 
Software has been developed to perform coreset selection
\citep{guo_deepcore_2022}, and coreset selection has been used in the
pre-training process of BERT models
\citep{%
  huang_pyramid-bert_2022,%
  attendu_nlu_2023%
}
as well as their fine-tuning
\citep{%
  nguyen_swift_2025,%
  sharma_towards_2025%
}, providing apparent compromise between performance and efficiency.
This has also been tested for LLMs \citep{
  san_joaquin_in2core_2024%
}.
An extensive analysis of coreset selection methods in deep learning however
found that ``although various methods have advantages in certain experimental
settings, random selection is still a strong baseline''
\citep[p.~181]{guo_deepcore_2022}, pointing to a potentially limited benefit of
complex methods given their cost, relative to simpler methods.

In NLP, diversity quantification is also used
for generative purposes
\citep{hu-etal-2019-large,zhou-lampouras-2021-informed-sampling,zhang-etal-2023-lingxi,yadav-etal-2024-explicit},
with an interest in the quality-diversity trade-off
\citep{zhang-etal-2021-trading}, for data augmentation
\citep{song-etal-2024-scaling}, or instruction tuning
\citep{yang-etal-2025-measuring}. In active learning, diversity is quantified
when
synthesizing
datasets from the ground up, rather than by sampling
\citep{shi-etal-2021-diversity,xia-etal-2024-hallucination}.

Approaches based on semantic structures present in data
\citep{oren-etal-2021-finding,gupta-etal-2022-structurally} or task-specific
objects such as in sentiment analysis where the sentiment of reviews can be
required for sampling \citep{jiang-etal-2023-large} have been studied. These
are motivated by task-specific rather than task-agnostic diversity, which can be for example based on lexical units.

We see mentions of distance-based sampling, as fostering higher distances
between studied categories may be a way to improve diversity. This is the case, for instance, in the k-means++ algorithm \citep{kim-2020-deep} or the previously mentioned
$k$-DPP sampling \citep{golobokov-etal-2022-deepgen}.

\citet[Algorithm~4]{chubarian-etal-2021-grouping} reference the algorithm by
\citet{lee_non-monotone_2009}
which iterates over a
dataset, trying to increase entropy at each data point,
by selecting it, unselecting it, or having it replace another selected point.

Given this rich bibliography related to diversity-driven sampling, we proceeded
to select algorithms which might match our framework and hypotheses
\ref{hyp:h1}-\ref{hyp:h2}.
The knapsack problem is NP-hard \citep{martello_new_2000} and recent findings
indicate that its lower bound is ``in subexponential and superpolynomial''
\citep[p.~11934]{zhang_lower_2025}, rendering it unfeasible for our needs.
$k$-DPP suffers a comparable issue.
To use $k$-DPP to sample a dataset, we
would need to produce a ``batch'' of the dataset size we want. However,
sampling takes at best polynomial time
to batch-size $k$ \citep{derezinski_exact_2019}. This is prohibitive if we want
a $k$ representing dataset size instead of batch size.
Approaches based on distances such as $k$-means++ can be filtered out,
as computing the matrix of distances is
quadratic.\footnote{Furthermore, using distances to measure diversity
(disparity) has issues, notably due to the high correlation between disparity
and vocabulary size \citep{estve-etal-2024-vector}.}
Diversity-driven approaches in decoding and in active learning are not
appropriate either, as we aim to algorithmically sample a dataset rather than
create one from scratch. The work of \citep{yang-etal-2025-measuring} has not been tested beyond 10,000 sentences, which is several orders of magnitude lower than our needs. 
Finally task-specific sampling does not fit our interest in task-agnostic
methods.

The approach used by \citet{chubarian-etal-2021-grouping} %
is adapted to scaling and can be used in
our study. We will compare it to two other algorithms proposed by us. One of them (§\ref{sec:our-algorithm-1}), like \citet{chubarian-etal-2021-grouping}, works on a local basis, i.e., actions are chosen only by looking at a limited subset of the
dataset. The other one accounts for the global nature of diversity and is based on gradient descent (§\ref{sec:our-algorithm-2}).
All three approaches will be compared to random sampling
\citep{jiang-etal-2023-large}.

\subsection{ModernBERT}\label{sec:modernbert}
\citet{warner-etal-2025-smarter}
proposed ModernBERT, which ports modern transformer improvements from auto-regressive models \cite{radford2018GPT} to masked architecture~\citep{devlin-etal-2019-bert}.
The main impact lies in the context extension from 512 to 8,192 tokens, based on a Rotary Positional Embedding \citep{SU2024127063}, also known as RoPE.
The GeGLU layers \citep{shazeer2020gluvariantsimprovetransformer} are preferred instead of MLP, which removes bias terms and adds layer normalization after embeddings. 

ModernBERT models provide new state-of-the-art performance on various tasks, such as Natural Language Understanding, Information Retrieval, Long-Context Text Retrieval, and Code Retrieval
\citep{warner-etal-2025-smarter}.

The original ModernBERT was trained on 2 trillion English tokens.
The original paper needed 1,879 and 4,076 hours of H100 computation respectively
to train
BASE and LARGE models.\footnote{``Training Time'' in
their Table~3, times GPU count.}

We aim to reduce these particularly high training costs by using a diversity-driven data sampling approach for the pre-training process.

\section{Sampling algorithms}\label{sec:sampling-algorithms}

We hereby describe sampling algorithms which we will run and compare. For each
algorithm, we shall test multiple parameters when possible. In the upcoming
descriptions, we use the following notations:
$v$ is the global vocabulary size, $p_i$ is the empirical probability of the $i$th
vocabulary entry, and $s$ is a sentence. \FULL{FULL} denotes the input dataset,
which comprises $n$ sentences. \DIVERSE{DIV} denotes the output dataset, which
comprises $n'$ sentences. Entropy is denoted as $H$.
As a \BASE{\textsc{Baseline}}, we perform random sampling with $q \in (0,1]$ as
the ratio of the sampled set to \FULL{FULL}.

\subsection{Patient add-only method}
\label{sec:our-algorithm-1}

Our first proposal is Algorithm~\ref{alg:hdc}. It
takes an initial dataset \BASE{BASE},
which is data the
Algorithm starts with to have a reasonable distribution balance-wise. It also
takes a maximum output size $S$, which can be disabled if $S = -1$.
In the internal loop
(lines~\ref{alg:foreach-document}-\ref{alg:foreach-document-end}), we consider
one candidate sentence $s$ from \EX{EXTENSION} at once. We filter and normalise
it (line~\ref{alg:norm}) to avoid artificial increase in
diversity.\footnote{(Telephone) numbers, HTML/XML tags, URLs, paths, emoticons,
punctuation series, phonetic characters, series of alphanumerical tokens, and
non-French characters are represented by placeholders, e.g., [NUMBER].} We
check if, when added to \DIVERSE{DIV$^*$}, $s$ increases the entropy $H$
(line~\ref{alg:test-increases-entropy}). If so, we check if this increase in
$H$ is higher than for a previously found sentence $best$
(line~\ref{alg:test-increases-entropy-more-than-current-best}). If so, $best$
becomes $s$ (line~\ref{alg:replace-d-with-n}).
Once sentences
improving the entropy $H$ have been found $e$ times, the best
is added to the final corpus \DIVERSE{DIV$^*$} (lines~\ref{alg:test-fulfilled-current-exhaustivity-level}-\ref{alg:append-to-corpus}) and we start looking for a new best sentence out of $e$ beneficial sentences (line~\ref{alg:reset-counter}).
If the
size of
\DIVERSE{DIV$^*$} is greater or equal to $S$, then early stopping is triggered, and
\DIVERSE{DIV$^*$} is returned
(line~\ref{alg:test-fulfilled-maximum-size}).

Variable $e$, for exhaustivity of search, tells us how many sentences
increasing diversity
to look
at before picking the optimal one to append to \DIVERSE{DIV$^*$}. The higher $e$,
the more each $best$ sentence may increase diversity. We use an array $E$ of
several exhaustivities, one per dataset traversal, sorted by decreasing values,
to have a deep search on the first traversal as the data has never been seen
before by the algorithm, and less so at each following traversal.
Without early stopping, we return \DIVERSE{DIV$^*$} when all exhaustivity levels
have been
used
(line~\ref{alg:return}). The algorithm has a
$O\left(\left\vert \text{\EX{EXTENSION}} \right\vert\right)$ complexity.
It was experimentally tested to create a large diverse automatically parsed corpus of French \citep{scholivet-etal-2025-selexini}.

\begin{algorithm}
\begin{algorithmic}[1]
  \Require \BASE{BASE} $\subset$ \FULL{FULL}, an initial dataset \label{alg:base}
    \Require \EX{EXTENSION} $=$ \FULL{FULL} $\setminus$ \BASE{BASE}, a dataset to select from
    \Require E, a decreasing array of exhaustivity search parameters (positive integers)
    \Require S, maximum size of resulting dataset ($-1$ if no maximum size)\label{alg:m}
    \State \DIVERSE{DIV$^*$} $\gets$ \BASE{BASE}, the final dataset \label{alg:init-W}
    \ForEach {e $\in$ E}, for each exhaustivity level \label{alg:foreach-exhaustivity-level}
        \State i $\gets$ 0, counter for current exhaustivity
        \State best $\gets \emptyset$, best sentence to append
        \ForEach {s $\in$ \EX{EXTENSION}} \label{alg:foreach-document}
            \State s $\gets$ {normalised}(s) \label{alg:norm}
            \If{H~(\DIVERSE{DIV$^*$}~$\cup$~s) > H~(\DIVERSE{DIV$^*$})} \label{alg:test-increases-entropy}
                \State i $\gets$ i + 1 \label{alg:increase-exhaustivity-counter}
                \If{H~(\DIVERSE{DIV$^*$}~$\cup$~s) > H~(\DIVERSE{DIV$^*$}~$\cup$~best)}\label{alg:test-increases-entropy-more-than-current-best}
                    \State best $\gets$ s \label{alg:replace-d-with-n}
                \EndIf
                \If{i = e} \label{alg:test-fulfilled-current-exhaustivity-level}
                    \State \DIVERSE{DIV$^*$} $\gets$ \DIVERSE{DIV$^*$} $\cup$~best \label{alg:append-to-corpus}
                    \State i $\gets$ 0 ; best $\gets \emptyset$ \label{alg:reset-counter}
                    \If{$-1<S\leq\left\vert{}\DIVERSE{\text{DIV}^*}\right\vert{}$} \label{alg:test-fulfilled-maximum-size}
                        \Return \DIVERSE{DIV$^*$} \label{alg:early-stopping-return}
                    \EndIf
                \EndIf
            \EndIf
        \EndFor \label{alg:foreach-document-end}
    \EndFor \label{alg:foreach-exhaustivity-level-end}    
    \State \Return \DIVERSE{DIV$^*$} \label{alg:return}
\end{algorithmic}
\caption{Algorithm by \citet{scholivet-etal-2025-selexini} to sample data while maximising entropy.\label{alg:hdc}}
\end{algorithm}

\subsection{Impatient add-remove-replace method}

\citet{lee_non-monotone_2009}
as presented by
\citet{chubarian-etal-2021-grouping}, also propose an algorithm based on
iterative dataset traversals. Unlike Algorithm~\ref{alg:hdc}, it has no memory of recent
favourable sentences. It starts with a \BASE{BASE} of just the sentence
with the highest entropy. It then traverses \FULL{FULL}. At each sentence $s$,
it assesses the impact on entropy for three different actions. If $s$ is not
already selected, it tests (1) adding $s$, or (2) replacing a random selected
sentence with $s$. If $s$ is already selected, it tests (3) unselecting $s$. If
the best action improves entropy by at least a margin (dependent on a parameter
$\epsilon$), this action is performed. If no update was done during a
traversal, the algorithm returns its working dataset as \DIVERSE{DIV}; otherwise
it starts another traversal. A pseudocode is given in
\citet[Algorithm~4]{chubarian-etal-2021-grouping}.

\subsection{Sampling by gradient descent}
\label{sec:our-algorithm-2}

We here sample by using a macroscopic view of the dataset. We do so with no \BASE{\textsc{BASE}}, using
gradient descent, where we learn for each sentence from \FULL{\textsc{FULL}} whether it should belong
in \DIVERSE{DIV}.
As a result,
each sentence is given a partial belonging $r \in [0, 1]$, which gets updated at
each gradient descent step.

We want to learn a vector of sentence belonging $b \in [0, 1]^{n}$. However,
learning directly this vector through gradient descent could give values
outside of $[0, 1]$. We thus instead learn a real row vector $a \in
\mathbb{R}^{n}$ which we then pass through the sigmoid function
$
    \text{sigm}\left(x\right)
    =
    \left(1 + e^{-x}\right)^{-1}
$
to obtain the $b \in [0, 1]^{n}$ row vector.
\begin{equation}
    \begin{bmatrix}
        a_1 & \cdots & a_n
    \end{bmatrix}\!\xrightarrow{\text{sigm}}\!\begin{bmatrix}
        b_1 & \cdots & b_n
    \end{bmatrix}
\end{equation}
We then multiply $b$ by $m$, the $n \times v$ matrix of the vocabulary for each
sentence (thus $m \in \left(\mathbb{N} \cup \left\{0\right\}\right)^{n \times v}$). This yields the $c = bm$ row vector of
word frequencies.
\begin{equation}
    \begin{bmatrix}
        b_1 & \cdots & b_n
    \end{bmatrix}\!\begin{bmatrix}
        m_{1,1} & \cdots & m_{1,v} \\
        \cdots & \cdots & \cdots \\
        m_{n,1} & \cdots & m_{n,v} \\
    \end{bmatrix}\!\rightarrow\!\begin{bmatrix}c_1 & \cdots & c_v\end{bmatrix}%
\end{equation}
We then use $c$ to compute vector $p$ of vocabulary probabilities
where
$
    p_i = c_i(\sum_{j=1}^{v}c_j)^{-1}
$
\begin{equation}
    \begin{bmatrix}c_1 & \cdots & c_v\end{bmatrix}\!\rightarrow\!\begin{bmatrix}p_1 & \cdots & p_v\end{bmatrix}
\end{equation}
We compute entropy $H$ over $p_1,\cdots,p_v$, which is the main element of the
loss function. If we negate it, gradient descent will try to increase entropy.
A shortcoming is that sentences may be selected with $b_i$ far from either $0$
or $1$, but we want in the end each sentence to be unambiguously close to $0$
(unselected) or $1$ (selected). To solve this, we add a penalty which increases
as $b_i$ gets distant from $0$ or $1$. This penalty is multiplied by $\alpha$
(default to 1000), and increases at each gradient descent step $t$, against the
maximum number of steps $T$.
\begin{equation}
    \mathcal{L} = (-H) + \frac{t}{T}\left(\alpha\frac{1}{n}\sum\limits_{i=1}^{n}b_i(1-b_i)\right) %
\end{equation}
At the end, we select sentences with $\left\lceil b_i \right\rfloor = 1$.

We have now presented the gradient descent approach, but there remains
to discuss the initialisation strategies for learned vector $a$.
We first test \textbf{random initialisation} where values of $a$ fall within
$[-1,1]$. This entails starting values in $b$ in the approximate range
$[0.27,0.73]$.

We also test \textbf{fixed initialisation} where all values of $a$ are initialised
at $\tanh\left(w\right)$ where $w \in \mathbb{R}$. If $w < 0$, then $b_i < 0.5$
(the start is pessimistic for each sentence), and conversely if $w >
0$, then $b_i > 0.5$ (the start is optimistic for each sentence).
Based on empirical results, Figure~\ref{fig:sampling-comparison} depicts eleven
values for $w$, in the range $[-0.25, 0.25]$ with a stepping of $0.05$. This
entails starting values in $b$ in the approximate range $[0.46,0.54]$.

We last test a \textbf{designed initialisation} based on
vocabulary
dissimilarity
between sentences, to benefit starting entropy. We first compute the $n \times n$
matrix $M = -\left(mm^{\intercal}\right)$ of dissimilarity between sentences.
We reduce a dimension to have a vector $d$, where $d_i =
n^{-1}\sum_{j=1}^{n}M_{i,j}$. We then apply normalisation, where $\mu$ is the
mean of $d$, and $\sigma$ its standard deviation, to obtain $a_i = 0.1 \times
\sigma^{-1} \left(d_i - \mu\right)$.

\subsection{Comparison of sampling methods}

\begin{figure*}%
    \resizebox{\textwidth}{!}{\input{img/sampling-plot.pgf}}
    \caption{Comparison of sampling algorithms run on UD
    v2.16 (French). \citet{lee_non-monotone_2009} have the best entropy,
    but
    ours-batched
    is just below,
    with only a third in $n'$. For
    ours-batched, the last value of $E$ is $20$, with
    each preceding value increasing by $10$ (the only exception is at
    $\left\vert E \right\vert = 1$ where the only value is
    $1$).\label{fig:sampling-comparison}}
\end{figure*}

We see in Figure~\ref{fig:sampling-comparison} the comparison of the algorithms
we described when sampling the union of French treebanks of Universal
Dependencies (UD) v2.16 \citep{nivre_universal_2020}. Sampling was done at the
sentence level. It is important to account for both
$n'$ (lower is better) and
$H$,
as comparing the diversity of datasets of
incommensurate
sizes is hazardous.

We first see that
non-random samplings
yield entropies markedly
higher than random sampling of
commensurate
output size $n'$, with the exception of
the gradient-based approach when using a fixed initialisation with a high $w$.
Among the studied methods, three reach
high entropies: \citet{lee_non-monotone_2009}, our gradient-based method (using a fixed initialisation with a low $w$), and our batch-based method (Algorithm~\ref{alg:hdc}).
We can assess which is best: gradient-based sampling has
high
memory requirements (it keeps the whole dataset in memory), and our batch-based method
yields
much smaller datasets
than that of
\citet{lee_non-monotone_2009}.
We thus
select Algorithm~\ref{alg:hdc}, which we denote by
``\DIVERSE{\textsc{Diverse}}'' for
our experiments.

\section{Pre-training protocol}\label{sec:experimental-protocol}

Our
protocol is as follows. We want to compare the
quality of encoders trained on \RANDOM{\textsc{Random}} and
\DIVERSE{\textsc{Diverse}} data. To do so, we use diversity-driven sampling to
create \DIVERSE{\textsc{Diverse}} datasets. For each \DIVERSE{\textsc{Diverse}}
dataset, we create a matching \RANDOM{\textsc{Random}} dataset of commensurate
size. For each pair of \DIVERSE{\textsc{Diverse}} and \RANDOM{\textsc{Random}}
datasets, we train two ModernBERT encoders, one using the former dataset, one
using the latter. We finish by fine-tuning on downstream tasks to
assess the quality of the \DIVERSE{\textsc{Diverse}} encoder against that of
the \RANDOM{\textsc{Random}} encoder. We give details in the following
subsections.

\subsection{Sampling}\label{sec:sampling}

We use Algorithm~\ref{alg:hdc}, 
explained in the previous section. We first select a \FULL{FULL} dataset. We
extract from it the \BASE{BASE}, which will be shared by all
\RANDOM{\textsc{Random}} and \DIVERSE{\textsc{Diverse}} models, while the rest
of the dataset is the \EX{EXTENSION}.\footnote{The use of a \BASE{BASE},
i.e. 
forcing the algorithm to start with already selected data, was assessed by
\citet{scholivet-etal-2025-selexini} to improve sampling for large datasets
containing noise.} In our case, the \BASE{BASE} is randomly-selected (approximately) 5\% of \FULL{FULL} (and thus the \EX{EXTENSION} is approximately 95\% of \FULL{FULL}).
For each \DIVERSE{\textsc{Diverse}} sampling configuration, the algorithm
selectively adds data from \EX{EXTENSION} to the \BASE{BASE} to create a
\DIVERSE{\textsc{Diverse}} dataset. For each \DIVERSE{\textsc{Diverse}} dataset
it generates, a \RANDOM{\textsc{Random}} dataset of commensurate size is created
by randomly adding data from the \EX{EXTENSION} to the \BASE{BASE}. The
pre-training data used to create the \FULL{FULL}, and in turn train the model,
is extracted from the French part of Wikipedia and OPUS
\citep{TIEDEMANN12.463}
corpora, which
enables us to propose some open-source models. The \FULL{FULL} is shuffled.
Each sampling takes less than 4 hours of CPU time.

\subsection{Pre-training datasets}\label{sec:results-sampling}

\begin{table}
\centering
\begin{tabular}{|l||r|r|r|} \hline
  Dataset & Words  & $H$' ($\uparrow$) & $H$ ($\uparrow$)   \\ \hline %
  \FULL{\textsc{Topline}}-2400M & 2~379M & $7.54$ & $7.55$ \\
 \hline
  \RANDOM{\textsc{Random}}-400M & 401M & $7.57$ & $7.59$ \\ %
  \DIVERSE{\textsc{Diverse}}-400M & 397M & $8.80$ & $8.53$      \\ \hline %
 \RANDOM{\textsc{Random}}-240M & 242M & $7.59$ & $7.60$ \\ %
  \DIVERSE{\textsc{Diverse}}-230M & 230M & $8.99$ & $8.45$    \\ \hline %
 \RANDOM{\textsc{Random}}-150M & 150M & $7.61$ & $7.61$ \\ %
  \DIVERSE{\textsc{Diverse}}-150M & 147M & $9.25$ & $8.21$     \\ \hline %
  \BASE{\textsc{Baseline}}-100M & 105M & N/A & $7.61$ \\ \hline
\end{tabular}
  \caption{Sampled datasets:
  part of \EX{EXTENSION} (with entropy $H'$) is appended to the \BASE{BASE} (the whole has entropy $H$).
  \DIVERSE{\textsc{Diverse}}-400M has E only at 1, \DIVERSE{\textsc{Diverse}}-230M from 16 to 4 (4 values; originally, it was planned to be from 50 to 20, but an error in settings made it run from 16 to 4) \DIVERSE{\textsc{Diverse}}-150M from 170 to 20 (16 values).\label{tab:encoders-training-data}}
\end{table}

The resulting datasets are presented in Table~\ref{tab:encoders-training-data}.
\BASE{\textsc{Baseline}}-100M is the \BASE{BASE} present in all datasets, with no
data from the \EX{EXTENSION}. Conversely, \FULL{\textsc{Topline}}-2400M uses
the whole \FULL{FULL}.
We see that each \DIVERSE{\textsc{Diverse}} dataset has an entropy notably higher
than its \RANDOM{\textsc{Random}} counterpart (i.e. with commensurate size). 
With an increasing $\left\vert E \right\vert$ (also due to
higher values in $E$),
the sampled data
from \EX{EXTENSION}
is more diverse, but also much smaller, meaning that once merged
with the \BASE{BASE}, diversity rankings are reversed.

\subsection{Pre-training process}\label{sec:pre-training}

ModernBERT has a native sequence length of 8,192 tokens and incorporates recent
architecture improvements, such as GeGLU layers, RoPE positional embeddings,
and alternating local-global attention. The vocabulary of the tokenizer is set
to 129K, and includes 1K unused tokens to support downstream applications. 

We train one base-size ModernBERT model
per dataset
from Table~\ref{tab:encoders-training-data}, using the
toolkit\footnote{\url{https://github.com/AnswerDotAI/ModernBERT}} given by
\citet{warner-etal-2025-smarter}. Each resulting model has 22 layers, a total parameter count of 200 million, and a hidden size
of 768 with a GLU expansion of 2,304. We assume a fixed pre-training compute budget of 483 hours of H100 per model. As a
comparison, the \FULL{\textsc{Topline}} using \FULL{FULL} took 1,775 hours of
GPU time (H100). 
The data, the pre-training recipe, and the models are openly available.

\subsection{\RANDOM{\textsc{Random}} and \DIVERSE{\textsc{Diverse}} encoders}\label{sec:results-pre-training}

\begin{figure}
\resizebox{\linewidth}{!}{\input{img/modernbert-fr-diversity.pgf}}
  \caption{Pre-training process for encoders. Loss is smoothed using a moving average (window size: $2 \times 10^5$, series sizes from $1.5$M to $6.3$M points) applied thrice.\label{fig:pre-training}}
\end{figure}

Pre-training on our different datasets has yielded loss profiles which are
mainly characterised by whether the training data was \RANDOM{\textsc{Random}}
or \DIVERSE{\textsc{Diverse}}
(Figure~\ref{fig:pre-training}).\footnote{Henceforth, the names of the datasets also denote the encoders trained upon.}
Looking at \RANDOM{\textsc{Random}} encoders,
more data means a
higher loss. This is also true for \DIVERSE{\textsc{Diverse}} encoders, but
their loss remains notably higher than their
\RANDOM{\textsc{Random}} counterpart.
About convergence speed, we see that \RANDOM{\textsc{Random}} encoders
all took approximately the same
amount of time.
\DIVERSE{\textsc{Diverse}} encoders had notably higher losses than their
respective \RANDOM{\textsc{Random}} counterparts.
The higher losses
are likely due to the higher difficulty of training due to diversity,
but conversely, it is possible that the encoder better models language and, especially, rare
phenomena. If so, we would expect \DIVERSE{\textsc{Diverse}} encoders to
perform better on downstream tasks.

\begin{table*}[t!]
  \begin{tabular}{|l|l|r|c|} \hline
    Task  (reference)  &   Domain                  &  Classes & Size (k)    \\      \hline \hline
        PAWS-X                  \citep{pawsx2019emnlp}   &  Paraphrase classification & $2$      & $49$ / $2.0$ / $2.0$         \\      \hline
        XNLI                      \citep{conneau2018xnli} &     NLI                  &  $3$    & $393$ / $5.4$ / $5.0$          \\      \hline
        MASSIVE Intent          \citep{fitzgerald-etal-2023-massive} &     Intent detection     &  $60$  & $11.5$ / $2.0$ / $2.9$            \\      \hline \hline
        WikiNER                 \citep{nothman2013wikiner} &      NER                  &       $7$   & $129$ / $0.5$ / $14.3$        \\      \hline
        MultiATIS++             \citep{upadhyay2018almost} &     POS-tagging          &       $131$   &  $37.0$ / $0.0$ / $7.8$                          \\      \hline
        Media (full)            \citep{CI_Bonneau-Maynard2006} &    NLU          &       $152$        &       $13.7$ / $1.3$ / $3.7$               \\      \hline
\end{tabular}
  \caption{Tasks used for fine-tuning. Size refers to the number of entries (by
  default, sentences), for TRAIN / DEV / TEST. For NER, the number of classes
  is after BIO expansion.\label{tab:task-presentation}}
\end{table*}

To evaluate this, and see the impact of our diversity sampling approach on the pre-training
data, we aim to perform several classical NLP tasks described in
Table~\ref{tab:task-presentation}: classification and sequence labelling tasks
(respectively first and second half of the Table). For multilingual
datasets, we use the French part. We selected tasks with an increasing number
of classes, assuming that it correlates with difficulty. 
It should be noted that the
MEDIA tasks have been considered the most challenging
ones by \cite{bechet2019benchmarking}.

\section{Results}\label{sec:results}

We used two models for the classification and sequence labelling tasks. 
Both use ModernBERT encoders, on which a width-preserving dense layer and a dense layer transforming to the number of classes are added as a decoder. 
The first uses simply the number of classes in the task. 
The second expands the list of labels to all the possible tags in the BIO scheme.

\subsection{Fine-tuning parameters}

Fine-tuning is performed in two ways. The first one fine-tunes both the encoder
and the head to see the maximum potential of the pipelines. The second one
freezes the encoder and trains the head, to assess encoder quality.
Hyperparameters are: learning rate is
$4 \times 10^{-4}$,
weight decay is
$1 \times 10^{-3}$,
max epochs is
$32$. The optimizer is Adam
($\beta_1=0.9$, $\beta_2 = 0.999$, $\epsilon = 1\times 10^{-8}$),
with an inverse square root scheduler ($15$\% warmup). Batch size
is $32$, evaluating every $64$ steps. We trigger an early stopping if $64$ train
losses are lower than
$1 \times 10^{-3}$,
or $32$ DEV scores not improving. Computation
is FP32.

\subsection{Fine-tuning results}\label{sec:results-fine-tuning}

\begin{table*}%
    \begin{center}
    \begin{tabular}{|l|r||c|c|c|c|c|c|}\hline
\multirow{2}{*}{Encoder}        &       \multirow{2}{*}{PTT}    &       \multicolumn{3}{|c|}{Classification} & \multicolumn{3}{|c|}{Sequence labelling} \\ \cline{3-8}
        &               &       PX      &       XNLI    &       AMI     &       WNER    &       MATIS   &       MDF\\ \hline
\FULL{\textsc{Topline}}-2400M & 1775h   &       84.7{\footnotesize{}$\pm$1.1}     &       75.9{\footnotesize{}$\pm$0.6}     &       81.7{\footnotesize{}$\pm$2.5}     &       89.8{\footnotesize{}$\pm$0.5}     &       92.5{\footnotesize{}$\pm$0.7}     &       82.6{\footnotesize{}$\pm$0.5} \\ \hline
\RANDOM{\textsc{Random}}-400M & 483h    &       85.9{\footnotesize{}$\pm$1.9}     &       75.1{\footnotesize{}$\pm$1.7}     &       \textbf{83.6{\footnotesize{}$\pm$0.7}}    &       90.0{\footnotesize{}$\pm$0.4}     &       92.1{\footnotesize{}$\pm$1.2}     &       \textbf{83.2{\footnotesize{}$\pm$0.4}} \\
\DIVERSE{\textsc{Diverse}}-400M & 483h  &       85.6{\footnotesize{}$\pm$1.5}     &       74.7{\footnotesize{}$\pm$0.6}     &       82.6{\footnotesize{}$\pm$1.0}     &       90.6{\footnotesize{}$\pm$0.9}     &       91.8{\footnotesize{}$\pm$1.8}     &       82.0{\footnotesize{}$\pm$1.1} \\ \cline{1-8}
\RANDOM{\textsc{Random}}-240M & 483h    &       82.6{\footnotesize{}$\pm$5.2}     &       \textbf{76.4{\footnotesize{}$\pm$1.0}}    &       82.5{\footnotesize{}$\pm$0.8}     &       90.1{\footnotesize{}$\pm$0.7}     &       92.8{\footnotesize{}$\pm$0.8}     &       82.7{\footnotesize{}$\pm$0.3} \\
\DIVERSE{\textsc{Diverse}}-230M & 483h  &       \textbf{86.3{\footnotesize{}$\pm$1.6}}    &       74.9{\footnotesize{}$\pm$1.8}     &       82.2{\footnotesize{}$\pm$1.5}     &       90.8{\footnotesize{}$\pm$0.4}     &       92.8{\footnotesize{}$\pm$0.8}     &       82.5{\footnotesize{}$\pm$0.2} \\ \cline{1-8}
\RANDOM{\textsc{Random}}-150M & 483h    &       84.5{\footnotesize{}$\pm$1.0}     &       72.3{\footnotesize{}$\pm$2.7}     &       82.0{\footnotesize{}$\pm$2.0}     &       89.5{\footnotesize{}$\pm$0.4}     &       92.5{\footnotesize{}$\pm$0.8}     &       82.4{\footnotesize{}$\pm$0.3} \\
\DIVERSE{\textsc{Diverse}}-150M & 483h  &       85.5{\footnotesize{}$\pm$0.7}     &       \textbf{75.0{\footnotesize{}$\pm$0.9}}    &       82.2{\footnotesize{}$\pm$1.9}     &       \textbf{90.5{\footnotesize{}$\pm$0.8}}    &       93.3{\footnotesize{}$\pm$0.3}     &       82.4{\footnotesize{}$\pm$0.4} \\ \cline{1-8}
\BASE{\textsc{Baseline}}-100M & 483h    &       83.2{\footnotesize{}$\pm$0.6}     &       74.7{\footnotesize{}$\pm$1.4}     &       83.0{\footnotesize{}$\pm$0.9}     &       89.5{\footnotesize{}$\pm$0.7}     &       92.5{\footnotesize{}$\pm$0.3}     &       78.0{\footnotesize{}$\pm$4.7} \\ \cline{1-8}
\end{tabular}
    \end{center}
    \caption{Encoder \& head fine-tuning. Evaluation on TEST. PTT is pre-training time. PX is
    PAWS-X,
    AMI is Amazon Massive Intent,
    WNER is WikiNER,
    MATIS is MultiATIS++,
    MDF is MEDIA (full).
    Mean $\pm$ standard deviation, across five
    seeds. Bold means $\Delta \geq 1$.\label{tab:score-summary-encoder-and-head}}
\end{table*}

\begin{table}%
    \begin{center}
    \begin{tabular}{|l|r||c|c|}\hline
Encoder  & PTT              &       MATIS   &       MDF\\ \hline
\FULL{\textsc{Topline}}-2400M & 1775h   &       83.2{\footnotesize{}$\pm$0.5}     &       58.9{\footnotesize{}$\pm$0.5} \\ \hline
\RANDOM{\textsc{Random}}-400M & 483h    &       84.0{\footnotesize{}$\pm$0.5}     &       \textbf{60.3{\footnotesize{}$\pm$0.3}} \\
\DIVERSE{\textsc{Diverse}}-400M & 483h  &       84.5{\footnotesize{}$\pm$0.3}     &       59.1{\footnotesize{}$\pm$0.8} \\ \cline{1-4}
\RANDOM{\textsc{Random}}-240M & 483h    &       84.4{\footnotesize{}$\pm$0.5}     &       59.7{\footnotesize{}$\pm$0.6} \\
\DIVERSE{\textsc{Diverse}}-230M & 483h  &       \textbf{86.4{\footnotesize{}$\pm$0.4}}    &       \textbf{61.7{\footnotesize{}$\pm$0.2}} \\ \cline{1-4}
\RANDOM{\textsc{Random}}-150M & 483h    &       80.2{\footnotesize{}$\pm$0.2}     &       51.9{\footnotesize{}$\pm$0.4} \\
\DIVERSE{\textsc{Diverse}}-150M & 483h  &       \textbf{84.3{\footnotesize{}$\pm$0.5}}    &       \textbf{\underline{61.5{\footnotesize{}$\pm$0.4}}} \\ \cline{1-4}
\BASE{\textsc{Baseline}}-100M & 483h    &       77.8{\footnotesize{}$\pm$0.2}     &       50.4{\footnotesize{}$\pm$0.7} \\ \cline{1-4}
\end{tabular}

    \end{center}
    \caption{Head-only fine-tuning. Evaluation on TEST. PTT is pre-training time. MDF is MEDIA
    (full), MATIS is MultiATIS++. Mean $\pm$ standard deviation, across five
    seeds. Bold means $\Delta \geq 1$, underline means $\Delta \geq
    5$.\label{tab:score-summary-head-only}}
    \vspace{-0.5cm}
\end{table}

Table~\ref{tab:score-summary-encoder-and-head} summarizes evaluations
of models with both encoders and heads fine-tuned,
on
the TEST of each task, where rows are grouped by
commensurate
pre-training
dataset sizes. A major observation is that both \RANDOM{\textsc{Random}} and \DIVERSE{\textsc{Diverse}} models yield a commensurate performance to \FULL{\textsc{Topline}}-2400M, despite reducing the pre-training dataset by up to 94\% and the training time by up to 73\%.\footnote{The \BASE{\textsc{Baseline}}, also performs comparatively well, except on MDF.} 
These results confirm former findings by \cite{guo_deepcore_2022} (§\ref{sec:sota-sampling}).
We find no clear signal, though, as to which of
\DIVERSE{\textsc{Diverse}} or \RANDOM{\textsc{Random}} models perform better. We
may hypothesize that, for these tasks, the model makes limited use of lexis, or
the lexical knowledge gained by \DIVERSE{\textsc{Diverse}} pre-training was not
impactful. 

Note, however, that our main objective is not to boost performances. We are instead particularly interested in the inherent quality of the encoders, independently of fine-tuning effects. Therefore, we perform the same evaluation, but with frozen encoders. This idea resembles probing classifiers, in which different layers of an encoder are used (with no fine-tuning) on input of simple classification tasks, to understand which knowledge they encode, rather than to optimize performances of these tasks.

In the experiment with frozen encoders, two tasks, MEDIA (full) and MultiATIS++, displayed marked improvements for \DIVERSE{\textsc{Diverse}} pre-training datasets over their \RANDOM{\textsc{Random}} counterpart; we display them in Table~\ref{tab:score-summary-head-only}.
For these tasks, relative to \BASE{\textsc{Baseline}}, the 50M tokens from the
\EX{EXTENSION} received by \RANDOM{\textsc{Random}}-150M and
\DIVERSE{\textsc{Diverse}}-150M had notably different consequences on
performance.
\RANDOM{\textsc{Random}}-150M saw improvements in average performance of respectively $1.5$ and $2.4$ points, while \DIVERSE{\textsc{Diverse}}-150M saw improvements in average performance of respectively $11.1$ and $6.5$ points, much higher than its counterpart.
This advantage of \DIVERSE{\textsc{Diverse}} over \RANDOM{\textsc{Random}}
pre-training data reduces between \DIVERSE{\textsc{Diverse}}-230M and
\RANDOM{\textsc{Random}}-240M, and vanishes between
\DIVERSE{\textsc{Diverse}}-400M and \RANDOM{\textsc{Random}}-400M.
It should also be noted that,
for these two tasks,
\DIVERSE{\textsc{Diverse}} encoders outperform the \FULL{\textsc{Topline}}, despite having considerably less pre-training data and lower number of pre-training steps (Figure~\ref{fig:pre-training}).
Looking back at the diversity of each pre-training dataset
(Table~\ref{tab:encoders-training-data}), we see that
this performance advantage is driven more by the diversity of the data
selected from the \EX{EXTENSION} than of the whole pre-training
dataset.
\DIVERSE{\textsc{Diverse}} pre-training datasets with fewer data from
\EX{EXTENSION} have sentences which individually contribute more to diversity,
as the sampler parameters were more selective, which may be what contributed to
the increased performance of these models, relative to their
\RANDOM{\textsc{Random}} counterparts.

The fact that this performance gain exists solely in head-only
fine-tuning is indicative of a fundamental difference between
\RANDOM{\textsc{Random}} and \DIVERSE{\textsc{Diverse}} encoders. Precisely, it
indicates that the \DIVERSE{\textsc{Diverse}} encoder contains more
task-important lexical knowledge. As, conversely, fine-tuning both the encoder
and the head yields commensurate performance for commensurately sized
pre-training datasets, we may hypothesize that fine-tuning
\RANDOM{\textsc{Random}} encoders compensates their prior lack of lexical
knowledge. This points to the fact that \DIVERSE{\textsc{Diverse}} encoders
have better potential -- at least for these tasks -- for data-constrained
scenarii.

We plot one of MEDIA's fine-tunings in
Figure~\ref{fig:media-speechbrain-full-y8-scenario2-cb0}. We see that
\DIVERSE{\textsc{Diversity}} provides an appreciable improvement in loss
and performance.
This
observation on MEDIA ports to MultiATIS++ in lesser proportions, but MEDIA is
further from being solved than MultiATIS++.

\begin{figure}%
\centering
  \resizebox{\linewidth}{!}{\input{img/loss_standard_media_speechbrain_full_y8_scenario2_cb0.pgf}}
\caption{Fine-tuning (head only) for MEDIA (full). Dashed thick lines
delimit
dataset traversals. Thick crosses denote maximum value.\label{fig:media-speechbrain-full-y8-scenario2-cb0}}
\vspace{-0.5cm}
\end{figure}

\section{Discussion\label{sec:discussion}}

We have shown that, (\ref{hyp:h1}) it is possible to efficiently create small
but very \DIVERSE{\textsc{Diverse}} pre-training datasets out of a large
dataset, and that (\ref{hyp:h2}) their use tends to have a neutral or positive
effect on encoder quality and performance, relative to \RANDOM{\textsc{Random}}
pre-training datasets of commensurate sizes, but also larger datasets.
Consequently,
it is possible to considerably trim pre-training datasets
while retaining performance. Precisely,
on the task benefitting the most from
lexical diversity (NLU),
for the head-only fine-tuning scenario,
diversity-driven sampling has allowed to reach the performance of an
encoder pre-trained on about 400M \RANDOM{\textsc{Random}} tokens, with only
150M \DIVERSE{\textsc{Diverse}} tokens.
Training time was also about a fourth that of the \FULL{\textsc{Topline}}. Future work includes
testing this process for auto-regressive models, and
investigating the impact of diversities other than lexical (e.g. syntactic).

\section{Limitations\label{sec:limitations}}

This study has the limitation of the potential correlation of lexical diversity
to other linguistic diversities, such as morphological, syntactic, or semantic
diversity. It may be that the encoders that benefitted from increased lexical
diversity were also impacted by changes these other linguistic diversities.

\section{Acknowledgements}
This work was performed using HPC resources from GENCI–IDRIS (Grant 2024-A0171013834). It was also supported by the CA21167 COST action UniDive, funded by COST (European Cooperation in Science and Technology).

\bibliography{bib/ecology.bib,bib/linguistic.bib,bib/stirling.bib,bib/information_theory.bib,bib/model.bib,bib/exp-012_diversity_diverse_2019-2024_ACL_2024-07-26.bib,bib/sampling.bib,bib/coreset.bib,latex/data.bib}

@inproceedings{indyk_composable_2014,
	address = {New York, NY, USA},
	series = {{PODS} '14},
	title = {Composable core-sets for diversity and coverage maximization},
	isbn = {978-1-4503-2375-8},
	url = {https://dl.acm.org/doi/10.1145/2594538.2594560},
	doi = {10.1145/2594538.2594560},
	urldate = {2026-04-13},
	booktitle = {Proceedings of the 33rd {ACM} {SIGMOD}-{SIGACT}-{SIGART} symposium on {Principles} of database systems},
	publisher = {Association for Computing Machinery},
	author = {Indyk, Piotr and Mahabadi, Sepideh and Mahdian, Mohammad and Mirrokni, Vahab S.},
	month = jun,
	year = {2014},
	pages = {100--108},
}

@inproceedings{sharma_towards_2025,
	address = {Suzhou, China},
	title = {Towards {Efficient} {FinBERT} via {Quantization} and {Coreset} for {Financial} {Sentiment} {Analysis}},
	url = {https://aclanthology.org/2025.finnlp-2.6/},
	doi = {10.18653/v1/2025.finnlp-2.6},
	urldate = {2026-04-13},
	booktitle = {Proceedings of {The} 10th {Workshop} on {Financial} {Technology} and {Natural} {Language} {Processing}},
	publisher = {Association for Computational Linguistics},
	author = {Sharma, Avinash Kumar and Hassan, Aisha Hamad and Shinde, Tushar},
	editor = {Chen, Chung-Chi and Winata, Genta Indra and Rawls, Stephen and Das, Anirban and Chen, Hsin-Hsi and Takamura, Hiroya},
	month = nov,
	year = {2025},
	pages = {70--74},
}

@article{agarwal_geometric_2005,
	title = {Geometric {Approximation} via {Coresets}},
	volume = {52},
	language = {en},
	number = {MSRI Publications},
	journal = {Combinatorial and Computational Geometry},
	publisher = {Cambridge University Press},
	author = {Agarwal, Pankaj K and Har-Peled, Sariel and Varadarajan, Kasturi R},
	editor = {Goodman, Jacob E. and Pach, Janos and Welzl, Emo},
	year = {2005},
}

@inproceedings{guo_deepcore_2022,
	address = {Cham},
	title = {{DeepCore}: {A} {Comprehensive} {Library} for {Coreset} {Selection} in {Deep} {Learning}},
	isbn = {978-3-031-12423-5},
	shorttitle = {{DeepCore}},
	doi = {10.1007/978-3-031-12423-5_14},
	language = {en},
	booktitle = {Database and {Expert} {Systems} {Applications}},
	publisher = {Springer International Publishing},
	author = {Guo, Chengcheng and Zhao, Bo and Bai, Yanbing},
	editor = {Strauss, Christine and Cuzzocrea, Alfredo and Kotsis, Gabriele and Tjoa, A. Min and Khalil, Ismail},
	year = {2022},
	pages = {181--195},
}

@inproceedings{huang_pyramid-bert_2022,
	address = {Dublin, Ireland},
	title = {Pyramid-{BERT}: {Reducing} {Complexity} via {Successive} {Core}-set based {Token} {Selection}},
	shorttitle = {Pyramid-{BERT}},
	url = {https://aclanthology.org/2022.acl-long.602/},
	doi = {10.18653/v1/2022.acl-long.602},
	urldate = {2026-04-13},
	booktitle = {Proceedings of the 60th {Annual} {Meeting} of the {Association} for {Computational} {Linguistics} ({Volume} 1: {Long} {Papers})},
	publisher = {Association for Computational Linguistics},
	author = {Huang, Xin and Khetan, Ashish and Bidart, Rene and Karnin, Zohar},
	editor = {Muresan, Smaranda and Nakov, Preslav and Villavicencio, Aline},
	month = may,
	year = {2022},
	pages = {8798--8817},
}

@inproceedings{attendu_nlu_2023,
	address = {Toronto, Canada (Hybrid)},
	title = {{NLU} on {Data} {Diets}: {Dynamic} {Data} {Subset} {Selection} for {NLP} {Classification} {Tasks}},
	shorttitle = {{NLU} on {Data} {Diets}},
	url = {https://aclanthology.org/2023.sustainlp-1.9},
	doi = {10.18653/v1/2023.sustainlp-1.9},
	language = {en},
	urldate = {2026-04-13},
	booktitle = {Proceedings of {The} {Fourth} {Workshop} on {Simple} and {Efficient} {Natural} {Language} {Processing} ({SustaiNLP})},
	publisher = {Association for Computational Linguistics},
	author = {Attendu, Jean-michel and Corbeil, Jean-philippe},
	year = {2023},
	pages = {129--146},
}

@inproceedings{san_joaquin_in2core_2024,
	address = {Miami, Florida, USA},
	title = {{In2Core}: {Leveraging} {Influence} {Functions} for {Coreset} {Selection} in {Instruction} {Finetuning} of {Large} {Language} {Models}},
	shorttitle = {{In2Core}},
	url = {https://aclanthology.org/2024.findings-emnlp.604/},
	doi = {10.18653/v1/2024.findings-emnlp.604},
	urldate = {2026-04-13},
	booktitle = {Findings of the {Association} for {Computational} {Linguistics}: {EMNLP} 2024},
	publisher = {Association for Computational Linguistics},
	author = {San Joaquin, Ayrton and Wang, Bin and Liu, Zhengyuan and Asher, Nicholas and Lim, Brian and Muller, Philippe and Chen, Nancy F.},
	editor = {Al-Onaizan, Yaser and Bansal, Mohit and Chen, Yun-Nung},
	month = nov,
	year = {2024},
	pages = {10324--10335},
}

@inproceedings{nguyen_swift_2025,
	address = {Abu Dhabi, UAE},
	title = {Swift {Cross}-{Dataset} {Pruning}: {Enhancing} {Fine}-{Tuning} {Efficiency} in {Natural} {Language} {Understanding}},
	shorttitle = {Swift {Cross}-{Dataset} {Pruning}},
	url = {https://aclanthology.org/2025.coling-main.49/},
	urldate = {2026-04-13},
	booktitle = {Proceedings of the 31st {International} {Conference} on {Computational} {Linguistics}},
	publisher = {Association for Computational Linguistics},
	author = {Nguyen, Binh-Nguyen and He, Yang},
	editor = {Rambow, Owen and Wanner, Leo and Apidianaki, Marianna and Al-Khalifa, Hend and Eugenio, Barbara Di and Schockaert, Steven},
	month = jan,
	year = {2025},
	pages = {726--739},
}

@inproceedings{perlitz_active_2023,
	address = {Singapore},
	title = {Active {Learning} for {Natural} {Language} {Generation}},
	url = {https://aclanthology.org/2023.emnlp-main.611},
	doi = {10.18653/v1/2023.emnlp-main.611},
	language = {en},
	urldate = {2026-04-13},
	booktitle = {Proceedings of the 2023 {Conference} on {Empirical} {Methods} in {Natural} {Language} {Processing}},
	publisher = {Association for Computational Linguistics},
	author = {Perlitz, Yotam and Gera, Ariel and Shmueli-Scheuer, Michal and Sheinwald, Dafna and Slonim, Noam and Ein-Dor, Liat},
	year = {2023},
	pages = {9862--9877},
}

@book{darwin_descent_1888,
	title = {The {Descent} of {Man},: {And} {Selection} in {Relation} to {Sex}},
	shorttitle = {The {Descent} of {Man},},
	language = {en},
	publisher = {John Murray, Albemarle Street.},
	author = {Darwin, Charles},
	year = {1888},
	note = {Google-Books-ID: NaPu24dY4iAC},
}

@article{hill_diversity_1973,
	title = {Diversity and {Evenness}: {A} {Unifying} {Notation} and {Its} {Consequences}},
	volume = {54},
	issn = {0012-9658},
	shorttitle = {Diversity and {Evenness}},
	url = {https://www.jstor.org/stable/1934352},
	doi = {10.2307/1934352},
	number = {2},
	urldate = {2023-10-06},
	journal = {Ecology},
	author = {Hill, Mark Oliver},
	year = {1973},
	note = {Number: 2
Publisher: Ecological Society of America},
	pages = {427--432},
}

@article{chao_unifying_2014,
	title = {Unifying {Species} {Diversity}, {Phylogenetic} {Diversity}, {Functional} {Diversity}, and {Related} {Similarity} and {Differentiation} {Measures} {Through} {Hill} {Numbers}},
	volume = {45},
	issn = {1543-592X},
	url = {https://www.jstor.org/stable/24810182},
	urldate = {2023-10-06},
	journal = {Annual Review of Ecology, Evolution, and Systematics},
	author = {Chao, Anne and Chiu, Chun-Huo and Jost, Lou},
	year = {2014},
	note = {Publisher: Annual Reviews},
	pages = {297--324},
}

@article{chiu_phylogenetic_2014,
        title = {Phylogenetic beta diversity, similarity, and differentiation measures based on {Hill} numbers},
        volume = {84},
        issn = {0012-9615, 1557-7015},
        url = {https://esajournals.onlinelibrary.wiley.com/doi/10.1890/12-0960.1},
        doi = {10.1890/12-0960.1},
        language = {en},
        number = {1},
        urldate = {2023-11-17},
        journal = {Ecological Monographs},
        author = {Chiu, Chun-Huo and Jost, Lou and Chao, Anne},
        month = feb,
        year = {2014},
        note = {Number: 1},
        pages = {21--44},
}

@article{sarkar_defining_2002,
        title = {Defining "{Biodiversity}"; {Assessing} {Biodiversity}},
        volume = {85},
        issn = {00269662},
        url = {http://www.jstor.org/stable/27903761},
        number = {1},
        urldate = {2025-12-05},
        journal = {The Monist},
        author = {Sarkar, Sahotra},
        year = {2002},
        note = {Publisher: Oxford University Press},
        pages = {131--155},
}

@article{patil_diversity_1982,
	title = {Diversity as a {Concept} and its {Measurement}},
	volume = {77},
	issn = {01621459},
	url = {http://www.jstor.org/stable/2287709},
	number = {379},
	urldate = {2023-07-20},
	journal = {Journal of the American Statistical Association},
	author = {Patil, Ganapati P. and Taillie, Charles},
	year = {1982},
	note = {Number: 379
Publisher: [American Statistical Association, Taylor \& Francis, Ltd.]},
	pages = {548--561},
}

@inproceedings{parrish-etal-2024-picture,
    title = "Is a picture of a bird a bird? A mixed-methods approach to understanding diverse human perspectives and ambiguity in machine vision models",
    author = "Parrish, Alicia  and
      Hao, Susan  and
      Laszlo, Sarah  and
      Aroyo, Lora",
    editor = "Abercrombie, Gavin  and
      Basile, Valerio  and
      Bernadi, Davide  and
      Dudy, Shiran  and
      Frenda, Simona  and
      Havens, Lucy  and
      Tonelli, Sara",
    booktitle = "Proceedings of the 3rd Workshop on Perspectivist Approaches to NLP (NLPerspectives) @ LREC-COLING 2024",
    month = may,
    year = "2024",
    address = "Torino, Italia",
    publisher = "ELRA and ICCL",
    url = "https://aclanthology.org/2024.nlperspectives-1.1",
    pages = "1--18",
}

@inproceedings{leeb-scholkopf-2024-diverse,
    title = "A diverse Multilingual News Headlines Dataset from around the World",
    author = {Leeb, Felix  and
      Sch{\"o}lkopf, Bernhard},
    editor = "Duh, Kevin  and
      Gomez, Helena  and
      Bethard, Steven",
    booktitle = "Proceedings of the 2024 Conference of the North American Chapter of the Association for Computational Linguistics: Human Language Technologies (Volume 2: Short Papers)",
    month = jun,
    year = "2024",
    address = "Mexico City, Mexico",
    publisher = "Association for Computational Linguistics",
    url = "https://aclanthology.org/2024.naacl-short.55",
    pages = "647--652",
}

@inproceedings{abdelkadir-etal-2024-diverse,
    title = "Diverse Perspectives, Divergent Models: Cross-Cultural Evaluation of Depression Detection on {T}witter",
    author = "Abdelkadir, Nuredin Ali  and
      Zhang, Charles  and
      Mayo, Ned  and
      Chancellor, Stevie",
    editor = "Duh, Kevin  and
      Gomez, Helena  and
      Bethard, Steven",
    booktitle = "Proceedings of the 2024 Conference of the North American Chapter of the Association for Computational Linguistics: Human Language Technologies (Volume 2: Short Papers)",
    month = jun,
    year = "2024",
    address = "Mexico City, Mexico",
    publisher = "Association for Computational Linguistics",
    url = "https://aclanthology.org/2024.naacl-short.58",
    pages = "672--680",
}

@inproceedings{xia-etal-2024-hallucination,
    title = "Hallucination Diversity-Aware Active Learning for Text Summarization",
    author = "Xia, Yu  and
      Liu, Xu  and
      Yu, Tong  and
      Kim, Sungchul  and
      Rossi, Ryan  and
      Rao, Anup  and
      Mai, Tung  and
      Li, Shuai",
    editor = "Duh, Kevin  and
      Gomez, Helena  and
      Bethard, Steven",
    booktitle = "Proceedings of the 2024 Conference of the North American Chapter of the Association for Computational Linguistics: Human Language Technologies (Volume 1: Long Papers)",
    month = jun,
    year = "2024",
    address = "Mexico City, Mexico",
    publisher = "Association for Computational Linguistics",
    url = "https://aclanthology.org/2024.naacl-long.479",
    pages = "8665--8677",
}

@inproceedings{yadav-etal-2024-explicit,
    title = "Explicit over Implict: Explicit Diversity Conditions for Effective Question Answer Generation",
    author = "Yadav, Vikas  and
      Kwon, Hyuk joon  and
      Srinivasan, Vijay  and
      Jin, Hongxia",
    editor = "Calzolari, Nicoletta  and
      Kan, Min-Yen  and
      Hoste, Veronique  and
      Lenci, Alessandro  and
      Sakti, Sakriani  and
      Xue, Nianwen",
    booktitle = "Proceedings of the 2024 Joint International Conference on Computational Linguistics, Language Resources and Evaluation (LREC-COLING 2024)",
    month = may,
    year = "2024",
    address = "Torino, Italia",
    publisher = "ELRA and ICCL",
    url = "https://aclanthology.org/2024.lrec-main.601",
    pages = "6876--6882",
}

@inproceedings{liu-etal-2024-prefix,
    title = "Prefix-diffusion: A Lightweight Diffusion Model for Diverse Image Captioning",
    author = "Liu, Guisheng  and
      Li, Yi  and
      Fei, Zhengcong  and
      Fu, Haiyan  and
      Luo, Xiangyang  and
      Guo, Yanqing",
    editor = "Calzolari, Nicoletta  and
      Kan, Min-Yen  and
      Hoste, Veronique  and
      Lenci, Alessandro  and
      Sakti, Sakriani  and
      Xue, Nianwen",
    booktitle = "Proceedings of the 2024 Joint International Conference on Computational Linguistics, Language Resources and Evaluation (LREC-COLING 2024)",
    month = may,
    year = "2024",
    address = "Torino, Italia",
    publisher = "ELRA and ICCL",
    url = "https://aclanthology.org/2024.lrec-main.1134",
    pages = "12954--12965",
}

@inproceedings{song-etal-2024-scaling,
    title = "Scaling Data Diversity for Fine-Tuning Language Models in Human Alignment",
    author = "Song, Feifan  and
      Yu, Bowen  and
      Lang, Hao  and
      Yu, Haiyang  and
      Huang, Fei  and
      Wang, Houfeng  and
      Li, Yongbin",
    editor = "Calzolari, Nicoletta  and
      Kan, Min-Yen  and
      Hoste, Veronique  and
      Lenci, Alessandro  and
      Sakti, Sakriani  and
      Xue, Nianwen",
    booktitle = "Proceedings of the 2024 Joint International Conference on Computational Linguistics, Language Resources and Evaluation (LREC-COLING 2024)",
    month = may,
    year = "2024",
    address = "Torino, Italia",
    publisher = "ELRA and ICCL",
    url = "https://aclanthology.org/2024.lrec-main.1251",
    pages = "14358--14369",
}

@inproceedings{samardzic-etal-2024-measure,
    title = "A Measure for Transparent Comparison of Linguistic Diversity in Multilingual {NLP} Data Sets",
    author = "Samardzic, Tanja  and
      Gutierrez, Ximena  and
      Bentz, Christian  and
      Moran, Steven  and
      Pelloni, Olga",
    editor = "Duh, Kevin  and
      Gomez, Helena  and
      Bethard, Steven",
    booktitle = "Findings of the Association for Computational Linguistics: NAACL 2024",
    month = jun,
    year = "2024",
    address = "Mexico City, Mexico",
    publisher = "Association for Computational Linguistics",
    url = "https://aclanthology.org/2024.findings-naacl.213",
    pages = "3367--3382",
}

@inproceedings{jiang-etal-2023-large,
    title = "Large-Scale and Multi-Perspective Opinion Summarization with Diverse Review Subsets",
    author = "Jiang, Han  and
      Wang, Rui  and
      Wei, Zhihua  and
      Li, Yu  and
      Wang, Xinpeng",
    editor = "Bouamor, Houda  and
      Pino, Juan  and
      Bali, Kalika",
    booktitle = "Findings of the Association for Computational Linguistics: EMNLP 2023",
    month = dec,
    year = "2023",
    address = "Singapore",
    publisher = "Association for Computational Linguistics",
    url = "https://aclanthology.org/2023.findings-emnlp.375",
    doi = "10.18653/v1/2023.findings-emnlp.375",
    pages = "5641--5656",
}

@inproceedings{gueuwou-etal-2023-jwsign,
    title = "{JWS}ign: A Highly Multilingual Corpus of {B}ible Translations for more Diversity in Sign Language Processing",
    author = {Gueuwou, Shester  and
      Siake, Sophie  and
      Leong, Colin  and
      M{\"u}ller, Mathias},
    editor = "Bouamor, Houda  and
      Pino, Juan  and
      Bali, Kalika",
    booktitle = "Findings of the Association for Computational Linguistics: EMNLP 2023",
    month = dec,
    year = "2023",
    address = "Singapore",
    publisher = "Association for Computational Linguistics",
    url = "https://aclanthology.org/2023.findings-emnlp.664",
    doi = "10.18653/v1/2023.findings-emnlp.664",
    pages = "9907--9927",
}

@inproceedings{fitzgerald-etal-2023-massive,
    title = "{MASSIVE}: A 1{M}-Example Multilingual Natural Language Understanding Dataset with 51 Typologically-Diverse Languages",
    author = "FitzGerald, Jack  and
      Hench, Christopher  and
      Peris, Charith  and
      Mackie, Scott  and
      Rottmann, Kay  and
      Sanchez, Ana  and
      Nash, Aaron  and
      Urbach, Liam  and
      Kakarala, Vishesh  and
      Singh, Richa  and
      Ranganath, Swetha  and
      Crist, Laurie  and
      Britan, Misha  and
      Leeuwis, Wouter  and
      Tur, Gokhan  and
      Natarajan, Prem",
    editor = "Rogers, Anna  and
      Boyd-Graber, Jordan  and
      Okazaki, Naoaki",
    booktitle = "Proceedings of the 61st Annual Meeting of the Association for Computational Linguistics (Volume 1: Long Papers)",
    month = jul,
    year = "2023",
    address = "Toronto, Canada",
    publisher = "Association for Computational Linguistics",
    url = "https://aclanthology.org/2023.acl-long.235",
    doi = "10.18653/v1/2023.acl-long.235",
    pages = "4277--4302",
}

@inproceedings{zhang-etal-2023-lingxi,
    title = "Lingxi: A Diversity-aware {C}hinese Modern Poetry Generation System",
    author = "Zhang, Xinran  and
      Sun, Maosong  and
      Liu, Jiafeng  and
      Li, Xiaobing",
    editor = "Bollegala, Danushka  and
      Huang, Ruihong  and
      Ritter, Alan",
    booktitle = "Proceedings of the 61st Annual Meeting of the Association for Computational Linguistics (Volume 3: System Demonstrations)",
    month = jul,
    year = "2023",
    address = "Toronto, Canada",
    publisher = "Association for Computational Linguistics",
    url = "https://aclanthology.org/2023.acl-demo.6",
    doi = "10.18653/v1/2023.acl-demo.6",
    pages = "63--75",
}

@inproceedings{pouran-ben-veyseh-etal-2022-minion,
    title = "{MINION}: a Large-Scale and Diverse Dataset for Multilingual Event Detection",
    author = "Pouran Ben Veyseh, Amir  and
      Nguyen, Minh Van  and
      Dernoncourt, Franck  and
      Nguyen, Thien",
    editor = "Carpuat, Marine  and
      de Marneffe, Marie-Catherine  and
      Meza Ruiz, Ivan Vladimir",
    booktitle = "Proceedings of the 2022 Conference of the North American Chapter of the Association for Computational Linguistics: Human Language Technologies",
    month = jul,
    year = "2022",
    address = "Seattle, United States",
    publisher = "Association for Computational Linguistics",
    url = "https://aclanthology.org/2022.naacl-main.166",
    doi = "10.18653/v1/2022.naacl-main.166",
    pages = "2286--2299",
}

@inproceedings{kumar-etal-2022-diversity,
    title = "{''}Diversity and Uncertainty in Moderation{''} are the Key to Data Selection for Multilingual Few-shot Transfer",
    author = "Kumar, Shanu  and
      Dandapat, Sandipan  and
      Choudhury, Monojit",
    editor = "Carpuat, Marine  and
      de Marneffe, Marie-Catherine  and
      Meza Ruiz, Ivan Vladimir",
    booktitle = "Findings of the Association for Computational Linguistics: NAACL 2022",
    month = jul,
    year = "2022",
    address = "Seattle, United States",
    publisher = "Association for Computational Linguistics",
    url = "https://aclanthology.org/2022.findings-naacl.78",
    doi = "10.18653/v1/2022.findings-naacl.78",
    pages = "1042--1055",
}

@inproceedings{gupta-etal-2022-structurally,
    title = "Structurally Diverse Sampling for Sample-Efficient Training and Comprehensive Evaluation",
    author = "Gupta, Shivanshu  and
      Singh, Sameer  and
      Gardner, Matt",
    editor = "Goldberg, Yoav  and
      Kozareva, Zornitsa  and
      Zhang, Yue",
    booktitle = "Findings of the Association for Computational Linguistics: EMNLP 2022",
    month = dec,
    year = "2022",
    address = "Abu Dhabi, United Arab Emirates",
    publisher = "Association for Computational Linguistics",
    url = "https://aclanthology.org/2022.findings-emnlp.365",
    doi = "10.18653/v1/2022.findings-emnlp.365",
    pages = "4966--4979",
}

@inproceedings{mohamed-etal-2022-artelingo,
    title = "{A}rt{EL}ingo: A Million Emotion Annotations of {W}iki{A}rt with Emphasis on Diversity over Language and Culture",
    author = "Mohamed, Youssef  and
      Abdelfattah, Mohamed  and
      Alhuwaider, Shyma  and
      Li, Feifan  and
      Zhang, Xiangliang  and
      Church, Kenneth  and
      Elhoseiny, Mohamed",
    editor = "Goldberg, Yoav  and
      Kozareva, Zornitsa  and
      Zhang, Yue",
    booktitle = "Proceedings of the 2022 Conference on Empirical Methods in Natural Language Processing",
    month = dec,
    year = "2022",
    address = "Abu Dhabi, United Arab Emirates",
    publisher = "Association for Computational Linguistics",
    url = "https://aclanthology.org/2022.emnlp-main.600",
    doi = "10.18653/v1/2022.emnlp-main.600",
    pages = "8770--8785",
}

@inproceedings{awasthi-etal-2022-diverse,
    title = "Diverse Parallel Data Synthesis for Cross-Database Adaptation of Text-to-{SQL} Parsers",
    author = "Awasthi, Abhijeet  and
      Sathe, Ashutosh  and
      Sarawagi, Sunita",
    editor = "Goldberg, Yoav  and
      Kozareva, Zornitsa  and
      Zhang, Yue",
    booktitle = "Proceedings of the 2022 Conference on Empirical Methods in Natural Language Processing",
    month = dec,
    year = "2022",
    address = "Abu Dhabi, United Arab Emirates",
    publisher = "Association for Computational Linguistics",
    url = "https://aclanthology.org/2022.emnlp-main.794",
    doi = "10.18653/v1/2022.emnlp-main.794",
    pages = "11548--11562",
}

@inproceedings{golobokov-etal-2022-deepgen,
    title = "{D}eep{G}en: Diverse Search Ad Generation and Real-Time Customization",
    author = "Golobokov, Konstantin  and
      Chai, Junyi  and
      Dong, Victor Ye  and
      Gu, Mandy  and
      Chi, Bingyu  and
      Cao, Jie  and
      Yan, Yulan  and
      Liu, Yi",
    editor = "Che, Wanxiang  and
      Shutova, Ekaterina",
    booktitle = "Proceedings of the 2022 Conference on Empirical Methods in Natural Language Processing: System Demonstrations",
    month = dec,
    year = "2022",
    address = "Abu Dhabi, UAE",
    publisher = "Association for Computational Linguistics",
    url = "https://aclanthology.org/2022.emnlp-demos.19",
    doi = "10.18653/v1/2022.emnlp-demos.19",
    pages = "191--199",
}

@inproceedings{burchell-birch-and-kenneth-heafield-2022-exploring,
    title = "Exploring diversity in back translation for low-resource machine translation",
    author = "Burchell, Laurie  and
      Birch, Alexandra  and
      Heafield, Kenneth",
    editor = "Cherry, Colin  and
      Fan, Angela  and
      Foster, George  and
      Haffari, Gholamreza (Reza)  and
      Khadivi, Shahram  and
      Peng, Nanyun (Violet)  and
      Ren, Xiang  and
      Shareghi, Ehsan  and
      Swayamdipta, Swabha",
    booktitle = "Proceedings of the Third Workshop on Deep Learning for Low-Resource Natural Language Processing",
    month = jul,
    year = "2022",
    address = "Hybrid",
    publisher = "Association for Computational Linguistics",
    url = "https://aclanthology.org/2022.deeplo-1.8",
    doi = "10.18653/v1/2022.deeplo-1.8",
    pages = "67--79",
}

@inproceedings{bella-etal-2022-language,
    title = "Language Diversity: Visible to Humans, Exploitable by Machines",
    author = "Bella, G{\'a}bor  and
      Byambadorj, Erdenebileg  and
      Chandrashekar, Yamini  and
      Batsuren, Khuyagbaatar  and
      Cheema, Danish  and
      Giunchiglia, Fausto",
    editor = "Basile, Valerio  and
      Kozareva, Zornitsa  and
      Stajner, Sanja",
    booktitle = "Proceedings of the 60th Annual Meeting of the Association for Computational Linguistics: System Demonstrations",
    month = may,
    year = "2022",
    address = "Dublin, Ireland",
    publisher = "Association for Computational Linguistics",
    url = "https://aclanthology.org/2022.acl-demo.15",
    doi = "10.18653/v1/2022.acl-demo.15",
    pages = "156--165",
}

@inproceedings{shi-etal-2021-diversity,
    title = "Diversity-Aware Batch Active Learning for Dependency Parsing",
    author = "Shi, Tianze  and
      Benton, Adrian  and
      Malioutov, Igor  and
      {\.I}rsoy, Ozan",
    editor = "Toutanova, Kristina  and
      Rumshisky, Anna  and
      Zettlemoyer, Luke  and
      Hakkani-Tur, Dilek  and
      Beltagy, Iz  and
      Bethard, Steven  and
      Cotterell, Ryan  and
      Chakraborty, Tanmoy  and
      Zhou, Yichao",
    booktitle = "Proceedings of the 2021 Conference of the North American Chapter of the Association for Computational Linguistics: Human Language Technologies",
    month = jun,
    year = "2021",
    address = "Online",
    publisher = "Association for Computational Linguistics",
    url = "https://aclanthology.org/2021.naacl-main.207",
    doi = "10.18653/v1/2021.naacl-main.207",
    pages = "2616--2626",
}

@inproceedings{chubarian-etal-2021-grouping,
    title = "Grouping Words with Semantic Diversity",
    author = "Chubarian, Karine  and
      Khan, Abdul Rafae  and
      Sidiropoulos, Anastasios  and
      Xu, Jia",
    editor = "Toutanova, Kristina  and
      Rumshisky, Anna  and
      Zettlemoyer, Luke  and
      Hakkani-Tur, Dilek  and
      Beltagy, Iz  and
      Bethard, Steven  and
      Cotterell, Ryan  and
      Chakraborty, Tanmoy  and
      Zhou, Yichao",
    booktitle = "Proceedings of the 2021 Conference of the North American Chapter of the Association for Computational Linguistics: Human Language Technologies",
    month = jun,
    year = "2021",
    address = "Online",
    publisher = "Association for Computational Linguistics",
    url = "https://aclanthology.org/2021.naacl-main.257",
    doi = "10.18653/v1/2021.naacl-main.257",
    pages = "3217--3228",
}

@inproceedings{zhang-etal-2021-trading,
    title = "Trading Off Diversity and Quality in Natural Language Generation",
    author = "Zhang, Hugh  and
      Duckworth, Daniel  and
      Ippolito, Daphne  and
      Neelakantan, Arvind",
    editor = "Belz, Anya  and
      Agarwal, Shubham  and
      Graham, Yvette  and
      Reiter, Ehud  and
      Shimorina, Anastasia",
    booktitle = "Proceedings of the Workshop on Human Evaluation of NLP Systems (HumEval)",
    month = apr,
    year = "2021",
    address = "Online",
    publisher = "Association for Computational Linguistics",
    url = "https://aclanthology.org/2021.humeval-1.3",
    pages = "25--33",
}

@inproceedings{zhou-lampouras-2021-informed-sampling,
    title = "Informed Sampling for Diversity in Concept-to-Text {NLG}",
    author = "Zhou, Giulio  and
      Lampouras, Gerasimos",
    editor = "Moens, Marie-Francine  and
      Huang, Xuanjing  and
      Specia, Lucia  and
      Yih, Scott Wen-tau",
    booktitle = "Findings of the Association for Computational Linguistics: EMNLP 2021",
    month = nov,
    year = "2021",
    address = "Punta Cana, Dominican Republic",
    publisher = "Association for Computational Linguistics",
    url = "https://aclanthology.org/2021.findings-emnlp.213",
    doi = "10.18653/v1/2021.findings-emnlp.213",
    pages = "2494--2509",
}

@inproceedings{bansal-etal-2021-diverse,
    title = "Diverse Distributions of Self-Supervised Tasks for Meta-Learning in {NLP}",
    author = "Bansal, Trapit  and
      Gunasekaran, Karthick Prasad  and
      Wang, Tong  and
      Munkhdalai, Tsendsuren  and
      McCallum, Andrew",
    editor = "Moens, Marie-Francine  and
      Huang, Xuanjing  and
      Specia, Lucia  and
      Yih, Scott Wen-tau",
    booktitle = "Proceedings of the 2021 Conference on Empirical Methods in Natural Language Processing",
    month = nov,
    year = "2021",
    address = "Online and Punta Cana, Dominican Republic",
    publisher = "Association for Computational Linguistics",
    url = "https://aclanthology.org/2021.emnlp-main.469",
    doi = "10.18653/v1/2021.emnlp-main.469",
    pages = "5812--5824",
}

@inproceedings{oren-etal-2021-finding,
    title = "Finding needles in a haystack: Sampling Structurally-diverse Training Sets from Synthetic Data for Compositional Generalization",
    author = "Oren, Inbar  and
      Herzig, Jonathan  and
      Berant, Jonathan",
    editor = "Moens, Marie-Francine  and
      Huang, Xuanjing  and
      Specia, Lucia  and
      Yih, Scott Wen-tau",
    booktitle = "Proceedings of the 2021 Conference on Empirical Methods in Natural Language Processing",
    month = nov,
    year = "2021",
    address = "Online and Punta Cana, Dominican Republic",
    publisher = "Association for Computational Linguistics",
    url = "https://aclanthology.org/2021.emnlp-main.843",
    doi = "10.18653/v1/2021.emnlp-main.843",
    pages = "10793--10809",
}

@inproceedings{kim-2020-deep,
    title = "Deep Active Learning for Sequence Labeling Based on Diversity and Uncertainty in Gradient",
    author = "Kim, Yekyung",
    editor = "Campbell, William M.  and
      Waibel, Alex  and
      Hakkani-Tur, Dilek  and
      Hazen, Timothy J.  and
      Kilgour, Kevin  and
      Cho, Eunah  and
      Kumar, Varun  and
      Glaude, Hadrien",
    booktitle = "Proceedings of the 2nd Workshop on Life-long Learning for Spoken Language Systems",
    month = dec,
    year = "2020",
    address = "Suzhou, China",
    publisher = "Association for Computational Linguistics",
    url = "https://aclanthology.org/2020.lifelongnlp-1.1",
    pages = "1--8",
}

@inproceedings{lee-etal-2020-generating,
    title = "Generating Diverse and Consistent {QA} pairs from Contexts with Information-Maximizing Hierarchical Conditional {VAE}s",
    author = "Lee, Dong Bok  and
      Lee, Seanie  and
      Jeong, Woo Tae  and
      Kim, Donghwan  and
      Hwang, Sung Ju",
    editor = "Jurafsky, Dan  and
      Chai, Joyce  and
      Schluter, Natalie  and
      Tetreault, Joel",
    booktitle = "Proceedings of the 58th Annual Meeting of the Association for Computational Linguistics",
    month = jul,
    year = "2020",
    address = "Online",
    publisher = "Association for Computational Linguistics",
    url = "https://aclanthology.org/2020.acl-main.20",
    doi = "10.18653/v1/2020.acl-main.20",
    pages = "208--224",
}

@inproceedings{stasaski-etal-2020-diverse,
    title = "More Diverse Dialogue Datasets via Diversity-Informed Data Collection",
    author = "Stasaski, Katherine  and
      Yang, Grace Hui  and
      Hearst, Marti A.",
    editor = "Jurafsky, Dan  and
      Chai, Joyce  and
      Schluter, Natalie  and
      Tetreault, Joel",
    booktitle = "Proceedings of the 58th Annual Meeting of the Association for Computational Linguistics",
    month = jul,
    year = "2020",
    address = "Online",
    publisher = "Association for Computational Linguistics",
    url = "https://aclanthology.org/2020.acl-main.446",
    doi = "10.18653/v1/2020.acl-main.446",
    pages = "4958--4968",
}

@inproceedings{hu-etal-2019-large,
    title = "Large-Scale, Diverse, Paraphrastic Bitexts via Sampling and Clustering",
    author = "Hu, J. Edward  and
      Singh, Abhinav  and
      Holzenberger, Nils  and
      Post, Matt  and
      Van Durme, Benjamin",
    editor = "Bansal, Mohit  and
      Villavicencio, Aline",
    booktitle = "Proceedings of the 23rd Conference on Computational Natural Language Learning (CoNLL)",
    month = nov,
    year = "2019",
    address = "Hong Kong, China",
    publisher = "Association for Computational Linguistics",
    url = "https://aclanthology.org/K19-1005",
    doi = "10.18653/v1/K19-1005",
    pages = "44--54",
}

@book{shannon_mathematical_1949,
        address = {Urbana},
        title = {A {Mathematical} {Theory} of {Communication}},
        language = {en},
        publisher = {University of Illinois Press},
        author = {Shannon, Claude Elwood and Weaver, Warren},
        year = {1949},
}

@article{ceriani_origins_2012,
        title = {The origins of the {Gini} index: extracts from {Variabilità} e {Mutabilità} (1912) by {Corrado} {Gini}},
        volume = {10},
        copyright = {http://www.springer.com/tdm},
        issn = {1569-1721, 1573-8701},
        shorttitle = {The origins of the {Gini} index},
        url = {http://link.springer.com/10.1007/s10888-011-9188-x},
        doi = {10.1007/s10888-011-9188-x},
        language = {en},
        number = {3},
        urldate = {2025-11-10},
        journal = {The Journal of Economic Inequality},
        author = {Ceriani, Lidia and Verme, Paolo},
        month = sep,
        year = {2012},
        pages = {421--443},
}

@article{wiener_ergodic_1939,
	title = {The ergodic theorem},
	volume = {5},
	issn = {0012-7094, 1547-7398},
	url = {https://projecteuclid.org/journals/duke-mathematical-journal/volume-5/issue-1/The-ergodic-theorem/10.1215/S0012-7094-39-00501-6.full},
	doi = {10.1215/S0012-7094-39-00501-6},
	number = {1},
	urldate = {2026-02-09},
	journal = {Duke Mathematical Journal},
	publisher = {Duke University Press},
	author = {Wiener, Norbert},
	month = mar,
	year = {1939},
	pages = {1--18},
}

@article{greenberg_measurement_1956,
	title = {The {Measurement} of {Linguistic} {Diversity}},
	volume = {32},
	issn = {00978507},
	url = {https://www.jstor.org/stable/410659?origin=crossref},
	doi = {10.2307/410659},
	language = {en},
	number = {1},
	urldate = {2023-07-20},
	journal = {Language},
	author = {Greenberg, Joseph Harold},
	month = jan,
	year = {1956},
	note = {Number: 1},
	pages = {109},
}

@InProceedings{bechet2019benchmarking,
  author      = {B{\'e}chet, Fr{\'e}d{\'e}ric and Raymond, Christian},
  title       = {{Benchmarking benchmarks: introducing new automatic indicators for benchmarking Spoken Language Understanding corpora}},
  booktitle   = {{InterSpeech}},
  year        = {2019},
  address     = {Graz, Austria},
  month       = Sep,
  hal_id      = {hal-02270633},
  hal_version = {v1},
  url         = {https://hal.science/hal-02270633},
}

@inproceedings{estve-etal-2024-vector,
    title = "Vector Spaces for Quantifying Disparity of Multiword Expressions in Annotated Text",
    author = "Est{\`e}ve, Louis  and
      Savary, Agata  and
      Lavergne, Thomas",
    editor = "Fu, Xiyan  and
      Fleisig, Eve",
    booktitle = "Proceedings of the 62nd Annual Meeting of the Association for Computational Linguistics (Volume 4: Student Research Workshop)",
    month = aug,
    year = "2024",
    address = "Bangkok, Thailand",
    publisher = "Association for Computational Linguistics",
    url = "https://aclanthology.org/2024.acl-srw.20/",
    doi = "10.18653/v1/2024.acl-srw.20",
    pages = "110--130",
    ISBN = "979-8-89176-097-4"
}

@article{harmon_index_2010,
	title = {The index of linguistic diversity: {A} new quantitative measure of trends in the status of the world's languages},
	volume = {4},
	issn = {1934-5275},
	shorttitle = {The index of linguistic diversity},
	url = {http://hdl.handle.net/10125/4474},
	language = {English},
	urldate = {2025-12-15},
	journal = {Language Documentation \& Conservation},
	author = {Harmon, David and Loh, Jonathan},
	month = sep,
	year = {2010},
	publisher = {University of Hawai'i Press},
	pages = {97--151},
}

@inproceedings{vaswani_attention_2017,
        address = {Red Hook, NY, USA},
        series = {{NIPS}'17},
        title = {Attention is all you need},
        isbn = {978-1-5108-6096-4},
        urldate = {2025-10-06},
        booktitle = {Proceedings of the 31st {International} {Conference} on {Neural} {Information} {Processing} {Systems}},
        publisher = {Curran Associates Inc.},
        author = {Vaswani, Ashish and Shazeer, Noam and Parmar, Niki and Uszkoreit, Jakob and Jones, Llion and Gomez, Aidan N. and Kaiser, Łukasz and Polosukhin, Illia},
        month = dec,
        year = {2017},
        pages = {6000--6010},
}

@misc{kaplan2020scalinglawsneurallanguage,
      title={Scaling Laws for Neural Language Models}, 
      author={Jared Kaplan and Sam McCandlish and Tom Henighan and Tom B. Brown and Benjamin Chess and Rewon Child and Scott Gray and Alec Radford and Jeffrey Wu and Dario Amodei},
      year={2020},
      eprint={2001.08361},
      archivePrefix={arXiv},
      primaryClass={cs.LG},
      url={https://arxiv.org/abs/2001.08361}, 
}

@inproceedings{strubell-etal-2019-energy,
    title = "Energy and Policy Considerations for Deep Learning in {NLP}",
    author = "Strubell, Emma  and
      Ganesh, Ananya  and
      McCallum, Andrew",
    editor = "Korhonen, Anna  and
      Traum, David  and
      M{\`a}rquez, Llu{\'i}s",
    booktitle = "Proceedings of the 57th Annual Meeting of the Association for Computational Linguistics",
    month = jul,
    year = "2019",
    address = "Florence, Italy",
    publisher = "Association for Computational Linguistics",
    url = "https://aclanthology.org/P19-1355/",
    doi = "10.18653/v1/P19-1355",
    pages = "3645--3650"
}

@misc{modernbert,
      title={Smarter, Better, Faster, Longer: A Modern Bidirectional Encoder for Fast, Memory Efficient, and Long Context Finetuning and Inference}, 
      author={Benjamin Warner and Antoine Chaffin and Benjamin Clavié and Orion Weller and Oskar Hallström and Said Taghadouini and Alexis Gallagher and Raja Biswas and Faisal Ladhak and Tom Aarsen and Nathan Cooper and Griffin Adams and Jeremy Howard and Iacopo Poli},
      year={2024},
      eprint={2412.13663},
      archivePrefix={arXiv},
      primaryClass={cs.CL},
      url={https://arxiv.org/abs/2412.13663}, 
}

@inproceedings{devlin-etal-2019-bert,
    title = "{BERT}: Pre-training of Deep Bidirectional Transformers for Language Understanding",
    author = "Devlin, Jacob  and
      Chang, Ming-Wei  and
      Lee, Kenton  and
      Toutanova, Kristina",
    editor = "Burstein, Jill  and
      Doran, Christy  and
      Solorio, Thamar",
    booktitle = "Proceedings of the 2019 Conference of the North {A}merican Chapter of the Association for Computational Linguistics: Human Language Technologies, Volume 1 (Long and Short Papers)",
    month = jun,
    year = "2019",
    address = "Minneapolis, Minnesota",
    publisher = "Association for Computational Linguistics",
    url = "https://aclanthology.org/N19-1423/",
    doi = "10.18653/v1/N19-1423",
    pages = "4171--4186"
}

@article{SU2024127063,
title = {RoFormer: Enhanced transformer with Rotary Position Embedding},
journal = {Neurocomputing},
volume = {568},
pages = {127063},
year = {2024},
issn = {0925-2312},
doi = {https://doi.org/10.1016/j.neucom.2023.127063},
url = {https://www.sciencedirect.com/science/article/pii/S0925231223011864},
author = {Jianlin Su and Murtadha Ahmed and Yu Lu and Shengfeng Pan and Wen Bo and Yunfeng Liu}
}

@misc{shazeer2020gluvariantsimprovetransformer,
      title={GLU Variants Improve Transformer}, 
      author={Noam Shazeer},
      year={2020},
      eprint={2002.05202},
      archivePrefix={arXiv},
      primaryClass={cs.LG},
      url={https://arxiv.org/abs/2002.05202}, 
}

@misc{esteve2026surveydiversityquantificationnatural,
      title={A survey of diversity quantification in natural language processing: The why, what, where and how}, 
      author={Louis Estève and Marie-Catherine de Marneffe and Nurit Melnik and Agata Savary and Olha Kanishcheva},
      year={2026},
      eprint={2507.20858},
      archivePrefix={arXiv},
      primaryClass={cs.CL},
      url={https://arxiv.org/abs/2507.20858}, 
}

@TechReport{radford2018GPT,
  author      = {Alec Radford and Karthik Narasimhan and Tim Salimans and Ilya Sutskever},
  institution = {Open AI},
  title       = {Improving Language Understanding by Generative Pre-Training},
  year        = {2018},
}

@inproceedings{warner-etal-2025-smarter,
    title = "Smarter, Better, Faster, Longer: A Modern Bidirectional Encoder for Fast, Memory Efficient, and Long Context Finetuning and Inference",
    author = {Warner, Benjamin  and
      Chaffin, Antoine  and
      Clavi{\'e}, Benjamin  and
      Weller, Orion  and
      Hallstr{\"o}m, Oskar  and
      Taghadouini, Said  and
      Gallagher, Alexis  and
      Biswas, Raja  and
      Ladhak, Faisal  and
      Aarsen, Tom  and
      Adams, Griffin Thomas  and
      Howard, Jeremy  and
      Poli, Iacopo},
    editor = "Che, Wanxiang  and
      Nabende, Joyce  and
      Shutova, Ekaterina  and
      Pilehvar, Mohammad Taher",
    booktitle = "Proceedings of the 63rd Annual Meeting of the Association for Computational Linguistics (Volume 1: Long Papers)",
    month = jul,
    year = "2025",
    address = "Vienna, Austria",
    publisher = "Association for Computational Linguistics",
    url = "https://aclanthology.org/2025.acl-long.127/",
    doi = "10.18653/v1/2025.acl-long.127",
    pages = "2526--2547",
    ISBN = "979-8-89176-251-0"
}

@article{cacchiani_knapsack_2022,
        title = {Knapsack problems — {An} overview of recent advances. {Part} {I}: {Single} knapsack problems},
        volume = {143},
        issn = {03050548},
        shorttitle = {Knapsack problems — {An} overview of recent advances. {Part} {I}},
        url = {https://linkinghub.elsevier.com/retrieve/pii/S0305054821003877},
        doi = {10.1016/j.cor.2021.105692},
        language = {en},
        urldate = {2025-02-14},
        journal = {Computers \& Operations Research},
        author = {Cacchiani, Valentina and Iori, Manuel and Locatelli, Alberto and Martello, Silvano},
        month = jul,
        year = {2022},
        pages = {105692},
}

@article{gong_diversity_2019,
        title = {Diversity in {Machine} {Learning}},
        volume = {7},
        copyright = {https://ieeexplore.ieee.org/Xplorehelp/downloads/license-information/OAPA.html},
        issn = {2169-3536},
        url = {https://ieeexplore.ieee.org/document/8717641/},
        doi = {10.1109/ACCESS.2019.2917620},
        language = {en},
        urldate = {2025-04-09},
        journal = {IEEE Access},
        author = {Gong, Zhiqiang and Zhong, Ping and Hu, Weidong},
        year = {2019},
        pages = {64323--64350},
}

@inproceedings{yang-etal-2025-measuring,
    title = "Measuring Data Diversity for Instruction Tuning: A Systematic Analysis and A Reliable Metric",
    author = "Yang, Yuming  and
      Nan, Yang  and
      Ye, Junjie  and
      Dou, Shihan  and
      Wang, Xiao  and
      Li, Shuo  and
      Lv, Huijie  and
      Gui, Tao  and
      Zhang, Qi  and
      Huang, Xuanjing",
    editor = "Che, Wanxiang  and
      Nabende, Joyce  and
      Shutova, Ekaterina  and
      Pilehvar, Mohammad Taher",
    booktitle = "Proceedings of the 63rd Annual Meeting of the Association for Computational Linguistics (Volume 1: Long Papers)",
    month = jul,
    year = "2025",
    address = "Vienna, Austria",
    publisher = "Association for Computational Linguistics",
    url = "https://aclanthology.org/2025.acl-long.908/",
    doi = "10.18653/v1/2025.acl-long.908",
    pages = "18530--18549",
    ISBN = "979-8-89176-251-0"
}

@inproceedings{scholivet-etal-2025-selexini,
    title = "{SELEXINI} {--} a large and diverse automatically parsed corpus of {F}rench",
    author = "Scholivet, Manon  and
      Savary, Agata  and
      Est{\`e}ve, Louis  and
      Candito, Marie  and
      Ramisch, Carlos",
    editor = "Sharoff, Serge  and
      Terryn, Ayla Rigouts  and
      Zweigenbaum, Pierre  and
      Rapp, Reinhard",
    booktitle = "Proceedings of the 18th Workshop on Building and Using Comparable Corpora (BUCC)",
    month = jan,
    year = "2025",
    address = "Abu Dhabi, UAE",
    publisher = "Association for Computational Linguistics",
    url = "https://aclanthology.org/2025.bucc-1.10/",
    pages = "83--98"
}

@inproceedings{derezinski_exact_2019,
        title = {Exact sampling of determinantal point processes with sublinear time preprocessing},
        volume = {32},
        url = {https://proceedings.neurips.cc/paper_files/paper/2019/hash/fa3060edb66e6ff4507886f9912e1ab9-Abstract.html},
        urldate = {2025-10-23},
        booktitle = {Advances in {Neural} {Information} {Processing} {Systems}},
        publisher = {Curran Associates, Inc.},
        author = {Derezinski, Michal and Calandriello, Daniele and Valko, Michal},
        year = {2019},
}

@inproceedings{yang_enhancing_2021,
        title = {Enhancing {Recommendation} {Diversity} {Using} {Determinantal} {Point} {Process} {Forward} {Inference} and {Backward} {Elimination}},
        url = {https://ieeexplore.ieee.org/document/9660543},
        doi = {10.1109/IC-NIDC54101.2021.9660543},
        urldate = {2025-10-23},
        booktitle = {2021 7th {IEEE} {International} {Conference} on {Network} {Intelligence} and {Digital} {Content} ({IC}-{NIDC})},
        author = {Yang, Xiaohan and Niu, Kun and Li, Xiao and Yu, Ruijie},
        month = nov,
        year = {2021},
        note = {ISSN: 2575-4955},
        pages = {56--60},
}

@article{hemmi_lazy_2022,
        title = {Lazy and {Fast} {Greedy} {MAP} {Inference} for {Determinantal} {Point} {Process}},
        volume = {35},
        url = {https://papers.nips.cc/paper_files/paper/2022/hash/127179162bfe4c422325ee7d05ad9cd8-Abstract-Conference.html},
        language = {en},
        urldate = {2025-12-18},
        journal = {Advances in Neural Information Processing Systems},
        author = {Hemmi, Shinichi and Oki, Taihei and Sakaue, Shinsaku and Fujii, Kaito and Iwata, Satoru},
        month = dec,
        year = {2022},
        pages = {2776--2789},
}

@article{zhang_lower_2025,
        title = {Lower bound of computational complexity of knapsack problems},
        volume = {10},
        copyright = {2025 The Author(s)},
        issn = {2473-6988},
        url = {http://www.aimspress.com/article/doi/10.3934/math.2025538},
        doi = {10.3934/math.2025538},
        language = {en},
        number = {5},
        urldate = {2025-10-27},
        journal = {AIMS Mathematics},
        author = {Zhang, Zhidong},
        year = {2025},
        note = {Cc\_license\_type: cc\_by
Primary\_atype: AIMS Mathematics
Subject\_term: Research article
Subject\_term\_id: Research article},
        pages = {11918--11938},
}

@article{martello_new_2000,
        title = {New trends in exact algorithms for the 0–1 knapsack problem},
        volume = {123},
        copyright = {https://www.elsevier.com/tdm/userlicense/1.0/},
        issn = {03772217},
        url = {https://linkinghub.elsevier.com/retrieve/pii/S037722179900260X},
        doi = {10.1016/S0377-2217(99)00260-X},
        language = {en},
        number = {2},
        urldate = {2025-10-27},
        journal = {European Journal of Operational Research},
        author = {Martello, Silvano and Pisinger, David and Toth, Paolo},
        month = jun,
        year = {2000},
        pages = {325--332},
}

@inproceedings{lee_non-monotone_2009,
        address = {New York, NY, USA},
        series = {{STOC} '09},
        title = {Non-monotone submodular maximization under matroid and knapsack constraints},
        isbn = {978-1-60558-506-2},
        url = {https://doi.org/10.1145/1536414.1536459},
        doi = {10.1145/1536414.1536459},
        urldate = {2025-12-18},
        booktitle = {Proceedings of the forty-first annual {ACM} symposium on {Theory} of computing},
        publisher = {Association for Computing Machinery},
        author = {Lee, Jon and Mirrokni, Vahab S. and Nagarajan, Viswanath and Sviridenko, Maxim},
        month = may,
        year = {2009},
        pages = {323--332},
}

@article{stirling_general_2007,
	title = {A general framework for analysing diversity in science, technology and society},
	volume = {4},
	url = {https://royalsocietypublishing.org/doi/10.1098/rsif.2007.0213},
	doi = {10.1098/rsif.2007.0213},
	number = {15},
	urldate = {2023-10-06},
	journal = {Journal of The Royal Society Interface},
	author = {Stirling, Andy},
	month = feb,
	year = {2007},
	note = {Number: 15
Publisher: Royal Society},
	pages = {707--719},
}

@article{stirling_diversity_1994,
	title = {Diversity and ignorance in electricity supply investment},
	volume = {22},
	copyright = {https://www.elsevier.com/tdm/userlicense/1.0/},
	issn = {03014215},
	url = {https://linkinghub.elsevier.com/retrieve/pii/0301421594901597},
	doi = {10.1016/0301-4215(94)90159-7},
	language = {en},
	number = {3},
	urldate = {2025-02-27},
	journal = {Energy Policy},
	author = {Stirling, Andrew},
	month = mar,
	year = {1994},
	pages = {195--216},
}

@phdthesis{stirling_power_1994,
	address = {Sussex},
	type = {{PhD}},
	title = {Power technology choice: putting the money where the mouth is?},
	url = {https://www.researchgate.net/publication/273138810_Power_Technology_Choice_Putting_the_money_where_the_mouth_is},
	language = {en},
	urldate = {2025-03-11},
	school = {University of Sussex},
	author = {Stirling, Andrew},
	year = {1994},
}

@article{ramaciotti_morales_measuring_2021,
        title = {Measuring diversity in heterogeneous information networks},
        volume = {859},
        url = {https://hal.science/hal-03608575},
        doi = {10.1016/j.tcs.2021.01.013},
        journal = {Theoretical Computer Science},
        author = {Ramaciotti Morales, Pedro and Lamarche-Perrin, Robin and Fournier-S'Niehotta, Raphaël and Poulain, Rémy and Tabourier, Lionel and Tarissan, Fabien},
        month = mar,
        year = {2021},
        note = {Publisher: Elsevier},
        pages = {80--115},
}

@InProceedings{CI_Bonneau-Maynard2006,
  author    = {Bonneau-Maynard, Hélène and Ayache, Christelle and Béchet, Frédéric and Denis, Alexandre and Kuhn, Anne and Lefevre, Fabrice and Mostefa, Djamel and Quignard, Mathieu and Rosset, Sophie and Servan, Christophe and Villaneau, Jeanne},
  title     = {{Results of the French Evalda-Media evaluation campaign for literal understanding}},
  booktitle = {{Proceedings of the fifth international conference on Language Resources and Evaluation (LREC 2006)}},
  year      = {2006},
  address   = {Genes, Italy},
  month     = {Mai},
}

\end{document}